\documentclass{article}

\PassOptionsToPackage{numbers, compress}{natbib}

\usepackage[main, final]{neurips_2026}

\usepackage[utf8]{inputenc} 
\usepackage[T1]{fontenc}    
\usepackage{hyperref}       
\usepackage{url}            
\usepackage{booktabs}       
\usepackage{amsfonts}       
\usepackage{nicefrac}       
\usepackage{microtype}      
\usepackage{xcolor}         
\usepackage{amsmath}
\usepackage{amssymb}
\usepackage{graphicx}
\usepackage{multirow}
\usepackage[table]{xcolor}
\usepackage{booktabs}
\usepackage{wrapfig}
\usepackage{algorithm}
\usepackage{algpseudocode}
\usepackage{amsmath}
\usepackage{longtable}
\usepackage{array}

\title{From Document Layout to Causal Topology: An End-to-End Architecture for Temporal Causal Reasoning}

\author{%
  \setcounter{footnote}{1}%
  Zhifeng Yang\(^{1}\), Wei Luo\(^{1}\), Yongjian Chen\(^{1}\), Shiqi Ou\(^{1,2}\thanks{Corresponding author.} \)\\
  \(^{1}\)School of Future Technology, South China University of Technology \\
  \(^{2}\)Guangdong Artificial Intelligence and Digital Economy Laboratory (Guangzhou) \\
  \texttt{ftyangzhifeng@mail.scut.edu.cn, 202364870991@mail.scut.edu.cn} \\
  \texttt{ftyjchen@mail.scut.edu.cn, sou@scut.edu.cn}
}

\begin{document}

\maketitle

\begin{abstract}
Large language models (LLMs) generally lack deep causal reasoning capabilities in temporal reasoning. Inspired by document layout reordering, this paper proposes a paradigm shift from discrete narratives to causal topologies. It constructs two architectures: a training-free neuro-symbolic framework, temporal causal reasoning agent (TCR-Agent), which achieves interpretable causal deduction and counterfactual truncation through external causal graph instantiation and a symbolic intervention engine; and an end-to-end model, temporal causal reasoning former (TCR-Former), which internalizes these mechanisms into the Transformer attention space via a causal-topological attention mask and temporal span biases. Experiments on the TempoCausal benchmark demonstrate that TCR-Agent comprehensively surpasses existing training-free baselines. At the same time, TCR-Former achieves an accuracy 16.5\% higher than that of state-of-the-art (SOTA) methods, outperforming large-scale models such as GPT-5.4 and DeepSeek-V3.2 with only approximately 8.6B parameters. Further analysis reveals a structure-reasoning disconnection phenomenon in existing methods: external symbolic constraints and chain-of-thought (CoT) reasoning can improve reasoning steps, yet fail to enhance deep causal deductive capability. By internalizing temporal causal reasoning capabilities into the model, TCR-Former effectively bridges this gap, providing an efficient and reliable end-to-end foundation for complex reasoning scenarios. The code and data can be obtained through https://github.com/Translab-SCUT/TCR-Former.
\end{abstract}

\section{Introduction}
Large language models (LLMs) have demonstrated strong generalization capabilities in tasks such as natural language understanding and instruction following \cite{R1,R2}. However, when tasks shift from time-independent semantic understanding to complex reasoning with temporal dependencies, existing models generally exhibit pronounced temporal blind spots \cite{R3,R4}. Studies indicate that LLMs perform poorly in temporal semantic parsing, long-context utilization, and complex event ordering, tending to rely on superficial textual order rather than genuine chronological order \cite{R5,R6}. To address this challenge, researchers have advanced the field along two dimensions: evaluation benchmarks and reasoning methods. On the evaluation side, early temporal question-answering benchmarks established foundational frameworks for task definition \cite{R7,R8,R9}. In contrast, subsequent evaluations have further revealed systematic deficiencies of LLMs in real-world temporality, multi-level dependencies, and cross-calendar reasoning \cite{R3,R5,R6,R10}. On the methodological side, recent work has partially improved performance in timeline ordering and explicit temporal question answering through techniques such as multi-hop temporal chain-of-thought (CoT) \cite{R11}, temporal graph intermediate representations \cite{R4}, and temporally-aware retrieval augmentation \cite{R12,R13}.

Nevertheless, the evolution of the aforementioned benchmarks and methods has primarily centered on the explicit modeling of event temporal relations, essentially remaining at the level of superficial timeline ordering \cite{R4,R56,R57} and generally lacking attention to the deep causal relationships underlying temporality. In real-world decision-making scenarios involving implicit temporal windows, multi-entity narratives, and counterfactual reasoning, models must not only sequence events chronologically but also capture the causal topology among events to support abductive and hypothetical inference. However, reasoning within structural causal models involves complex operations such as abduction, intervention, and counterfactuals, which cannot be readily achieved through simple event ordering \cite{R14} or flat retrieval \cite{R15}. Some efforts have attempted to enhance reasoning reliability through causalized CoT \cite{R16,R17}. Still, these methods generally impose causal structure by externally controlling the reasoning process and do not genuinely internalize causal topology awareness and intervention mechanisms into the model's forward computation, thus remaining constrained in cross-scenario generalization and reasoning efficiency \cite{R57}. This predicament reveals that merely superimposing causal structure through external symbolic constraints does not address the essence of the problem. The crux of temporal causal reasoning lies in internalizing the topology-aware capacity for temporal causation into the model's reasoning process.

Inspired by visual document understanding, we observe a profound computational isomorphism between complex temporal causal reconstruction and layout structuring tasks. ROAP \cite{R20} and DeepSeek-OCR 2 \cite{R21} reorganize discrete, disordered visual blocks in two-dimensional space into logically ordered one-dimensional sequences through reading-order-aware biases and visual causal flow mechanisms. The core challenge faced by temporal reasoning is highly analogous; both essentially require reconstructing discrete, disordered perceptual fragments into a structured logical topology. Based on this insight, this paper aims to transfer the logic of layout reordering to temporal causal reconstruction by modeling the causal topology among events and internalizing causal topology awareness and temporal priors into the model's reasoning process and attention mechanisms.

Motivated by the above, we propose the temporal causal reasoning agent (TCR-Agent) framework, which explicitly reconstructs the causal topology in complex narratives through an external causal graph and a symbolic intervention engine. To further enhance end-to-end reasoning efficiency and integrate temporal causal priors into the underlying model representations, we extend this framework by proposing the temporal causal reasoning former (TCR-Former), which employs a causal-topological attention mask, explicit temporal span biases, and a native counterfactual intervention mechanism to achieve end-to-end temporal causal topology awareness.

The main contributions of this paper are as follows.

\begin{itemize}
    \item We construct the TempoCausal-Bench to address the lack of deep causal reasoning evaluation in existing benchmarks.

    \item We propose the TCR-Agent framework, a training-free, model-agnostic external causal graph and symbolic intervention scheme compatible with various open-source and proprietary LLMs, providing interpretable and intervenable system-level support for complex temporal causal reasoning.

    \item We propose the TCR-Former model, an end-to-end architecture that internalizes causal topology awareness and intervention mechanisms into Transformer forward computation, providing an efficient and reliable temporal causal reasoning backbone for complex reasoning scenarios.
\end{itemize}

\section{TCR-Agent Algorithm}
TCR-Agent adopts a neuro-symbolic reasoning architecture and the core of TCR-Agent lies in leveraging an external graph memory mechanism to reconstruct the latent causal structure, enabling a pre-trained LLM to perform verifiable counterfactual reasoning and multi-hop temporal deduction without modifying its internal parameters. The framework formally integrates the structural causal model (SCM) and the potential outcome framework (POF). SCM provides a formal representation and operations for the causal topology, while POF is utilized to evaluate evolutionary outcomes under counterfactual settings. The core algorithmic flow is presented in Algorithm \ref{alg:1} (Appendix \ref{app: TCR-Agent and TCR-Former Algorithm}).

\subsection{Causal Graph Extraction}

This module aims to reconstruct the causal topology among events from unstructured text explicitly. We design a query-aware spatiotemporal extractor $E_\theta$ and a symbolic graph constructor $B$ to work in concert, achieving end-to-end construction of a structured causal directed acyclic graph (DAG) from raw text. Temporal question answering typically involves one or more text segments describing the background of a question. Assume it is a sequence of $N_c$ text chunks, $C=\{c_1, \cdots, c_{N_c}\}$. Let $P$ denote the user query prompt, composed of a temporal question and other relevant hints, such as a CoT for solving the question. The spatiotemporal extractor $E_\theta$ takes $C$ and $P$ as input and extracts a causal graph by invoking the LLM. The extraction process is shown in Equations (\ref{eq:1}) and (\ref{eq:2}).

\begin{align}
(S, G_{\mathrm{raw}}) &= E_\theta(C, P) \triangleq \mathrm{LLM}_\theta(C, P) \label{eq:1} \\
G_{\mathrm{raw}} &= (\mathcal{V}_{\mathrm{raw}}, \mathcal{E}_{\mathrm{raw}}) \label{eq:2}
\end{align}

where, $S$ records intermediate reasoning steps such as date differences, quantity accumulation, and temporal boundary determination to suppress information decay and hallucinations during long-context parsing. $G_{\mathrm{raw}}$ denotes the raw graph data, with $\mathcal{V}_{\mathrm{raw}}$ being the set of extracted event nodes and $\mathcal{E}_{\mathrm{raw}}$ the set of causal edges among these nodes.

Subsequently, the symbolic graph constructor $B$ instantiates $G_{\mathrm{raw}}$ into a directed acyclic graph $G = (\mathcal{V}, \mathcal{E})$. For each node $v_i \in \mathcal{V}$, its original temporal anchor $t_i$ is preserved, and its textual description and quantitative data are fused into a unified semantic representation, as formulated in Equation (\ref{eq:3}).

\begin{equation}
\tilde{v}_i = \mathrm{desc}_i \oplus \mathrm{Annotate}(d_i), \forall v_i \in \mathcal{V} \label{eq:3}
\end{equation}

where $\mathrm{desc}_i$ is the textual description of the node $v_i$, $d_i$ is the quantitative numerical field, $\mathrm{Annotate}(\cdot)$ denotes a function that formats numerical data into explicit annotation text, and $\oplus$ denotes string concatenation. Consequently, each node simultaneously carries the temporal anchor $t_i$ and the enriched semantic representation $\tilde{v}_i$, forming a three-dimensional representation space that fuses temporal, semantic, and numerical information.

For each directed edge $e_{ij} \in \mathcal{E}$, the associated attributes include the source node $src_i$, the target node $tgt_j$, the relation type $type_{ij}$, and the reason $reason_{ij}$, where $reason_{ij}$ records the causal derivation basis from the source event to the target event. During instantiation, the graph constructor $B$ resolves potential contradictions among node timestamps via topological sorting, enforces the acyclicity constraint, and ultimately outputs a strictly structured logical trajectory $G_{\mathrm{logic}}$.

To distinguish between causal and non-causal relationships among events, we categorize events in complex temporal question answering scenarios into three types and apply differentiated processing strategies:

\textbf{(1) Irrelevant Events:} Events with no logical intersection in semantics or time. The system directly filters these out during the causal graph extraction phase.

\textbf{(2) Co-occurring Events:} Events that merely overlap or follow a sequential order in time, but lack substantial causal drivers. The system retains only the independent states and temporal anchors of these nodes in the causal graph, without establishing any causal directed edge connections.

\textbf{(3) Causal Events:} Events with explicit causal dependencies. The system not only establishes directed edge connections but also generates attribute fields such as $reason$ to accurately characterize and record the specific causal derivation basis from the source event to the target event.

\subsection{Counterfactual Intervention}
This module aims to support what-if reasoning by performing symbolic-level structural perturbations to truncate and redirect causal chains. To this end, we design a graph intervention engine $\mathcal{I}$. Semantically, this intervention operation is equivalent to the do-operator in SCM, cutting off incoming edges to specific variables via counterfactual truncation and observing system changes. The intervention operation is triggered by counterfactual assumptions in $P$. When $P$ contains counterfactual statements such as "if an event had not occurred," the LLM parses the intervention parameters $\psi = (v_k, action)$, where $v_k$ is the target node and $action \in \{\mathrm{remove}, \mathrm{cut\_edges}\}$ specifies the type of intervention. Given the intervention parameters $\psi$, $\mathcal{I}$ applies structural perturbations to the current causal graph $G = (\mathcal{V}, \mathcal{E})$ to produce a modified causal graph $G'=(\mathcal{V}', \mathcal{E}')$, as computed in Equation (\ref{eq:4}).

\begin{equation}
G' = \mathcal{I}(G, \psi) = 
\begin{cases} 
G \setminus \{v_k\} \cup \mathcal{E}_{\mathrm{preserve}}(v_k), & \text{if } action = \mathrm{remove} \\ 
G \setminus \{(v_k, u) | (v_k, u) \in \mathcal{E}\}, & \text{if } action = \mathrm{cut\_edges} 
\end{cases}
\label{eq:4}
\end{equation}

where, $u$ denotes the target node of a directed edge originating from $v_k$ in the causal graph. When $action = \mathrm{remove}$, the target node $v_k$ is removed from the graph, while certain adjacent edges are explicitly preserved via $\mathcal{E}_{\mathrm{preserve}}(v_k)$ to maintain the necessary connectivity of the narrative context. When $action = \mathrm{cut\_edges}$, all outgoing edges from $v_k$ are deleted while the node itself is retained. The perturbed causal graph $G'$ is linearized via a depth-first traversal into a topological description string $D(G')$, which explicitly encodes the temporal gaps, semantic descriptions, and causal justifications between event nodes.

\subsection{Dual-Track Reasoning}
This module aims to constrain autoregressive generation into a verifiable symbolic execution process through serialized cognitive procedures and explicit invariant checking. The core reasoning is executed by a dual-track reasoner $R$, which integrates the unstructured raw text $C$ and the structured logical trajectory $D(G')$. To regulate the LLM's reasoning process, we constrain the generation trajectory using serialized tags organized into three parts: checking, reasoning, and reflection. Their specific contents are as follows. 

\textbf{(1) Checking:} Describes $D(G')$ in natural language, and based on $C$ checks the temporal computation correctness and causal derivation consistency of $D(G')$, and if incorrect, corrects $D(G')$. 

\textbf{(2) Reasoning:} Performs reasoning according to $C$, $P$, and $D(G')$, and returns Unknown if the information is insufficient. 

\textbf{(3) Reflection:} Reflects on whether the answer is correct, and if the answer is inconsistent with $C$ and $D(G')$, corrects it and returns the correct answer.

\section{TCR-Former Architecture}
TCR-Former embeds the temporal causal modeling logic of the TCR-Agent paradigm into the Transformer's forward computation pipeline. TCR-Former consists of a temporal causal encoder and a reasoning decoder. The encoder extracts high-dimensional causal latent variables from fragmented text, and the decoder feeds these latent variables, along with the original factual information and user query instructions, into the autoregressive generation process. The architecture of the TCR-Former model is illustrated in Fig. \ref{fig:1}. The training strategy of TCR-Former is detailed in Appendix \ref{app:TCR-Former Training Strategy}.

\begin{figure}[t]
\centering
\includegraphics[width=14cm]{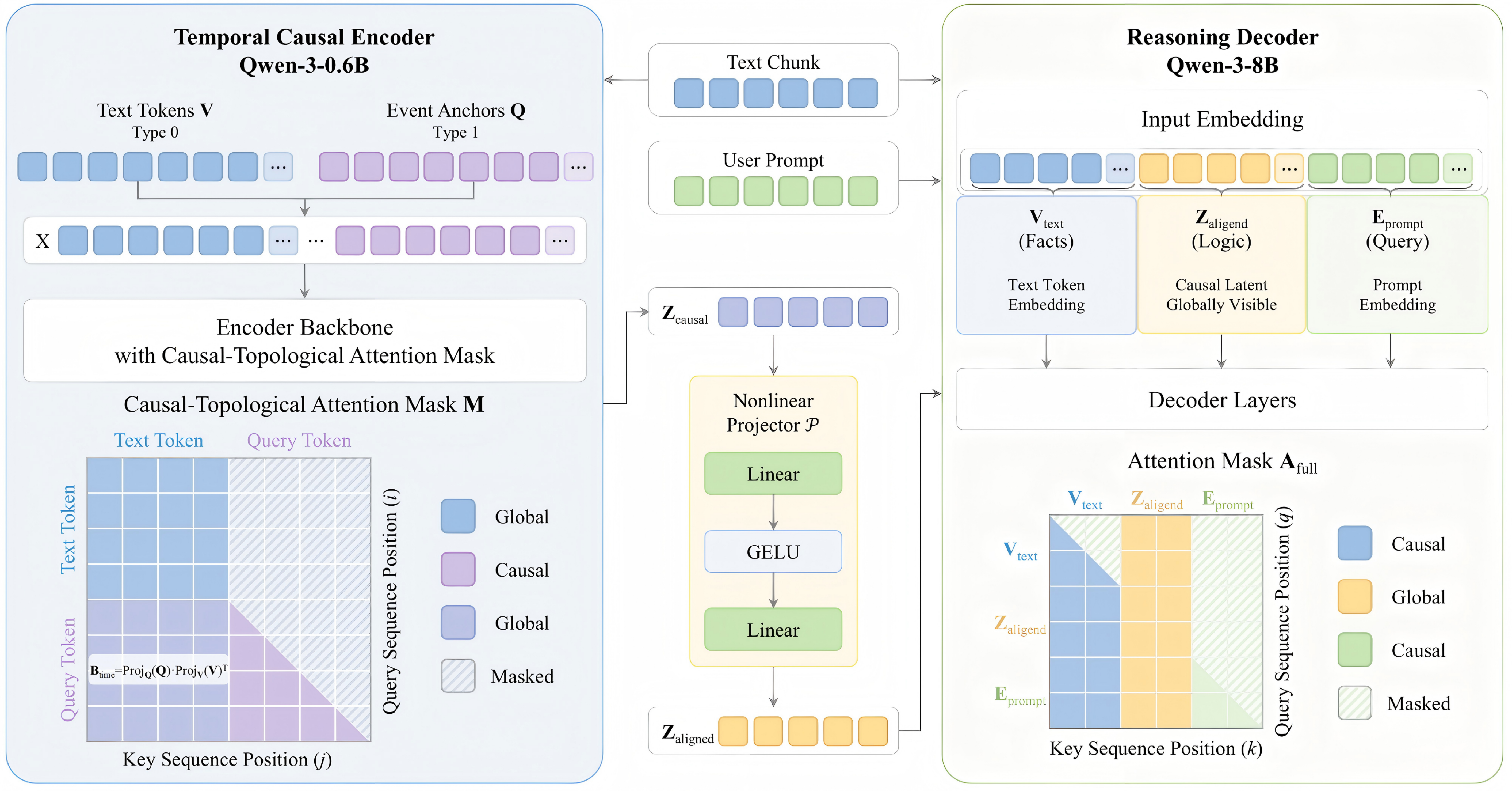}
\caption{TCR-Former architecture.}
\label{fig:1}
\end{figure}

\subsection{Temporal Causal Encoder}

The temporal causal encoder uses Qwen-3-0.6B as its backbone, aiming to extract structured temporal causal latent variables from unstructured text fragments.

\subsubsection{Input Sequence Processing}

For a sequence of $N_c$ text chunks $C=\{c_1, \cdots, c_{N_c}\}$, the model first converts them into continuous text token representations $\mathbf{V} \in \mathbb{R}^{B \times L_c \times d_{\mathrm{enc}}}$ through a word embedding layer, where $B$ is the batch size, $L_c$ is the total length of the concatenated text tokens, and $d_{\mathrm{enc}}$ is the encoder hidden dimension. To discover the implicit causal structure within the text, we introduce a set of $N$ learnable event anchors as a causal query token embedding matrix $\mathbf{Q}=\{\mathbf{q}_1, \cdots, \mathbf{q}_N\}$. Each event anchor $\mathbf{q}_k$ is used to query and aggregate a potential causal event from the text. In experiments, we set $N = 128$, which exceeds the maximum possible number of events in any sample, and the surplus slots are set to an empty event state during training. $\mathbf{q}_k$ consists of two components, as shown in Equation (\ref{eq:5}).

\begin{equation}
\mathbf{q}_k = \mathbf{E}_{\mathrm{anchor}}(k) + \mathbf{E}_{\mathrm{span}}(k), \quad k \in [1, N]
\label{eq:5}
\end{equation}

where, $\mathbf{E}_{\mathrm{anchor}} \in \mathbb{R}^{N \times d_{\mathrm{enc}}}$ is a position-independent base anchor embedding that provides an independent basis vector for each potential event slot. $\mathbf{E}_{\mathrm{span}} \in \mathbb{R}^{N \times d_{\mathrm{enc}}}$ is a learnable time span bias embedding, which endows each event slot with a different temporal domain prior, for example, making certain anchors more inclined to attend to short-term events while others to long-range events. Both $\mathbf{E}_{\mathrm{anchor}}$ and $\mathbf{E}_{\mathrm{span}}$ are trainable parameters.

The text tokens and query tokens are finally concatenated into a one-dimensional input sequence $\mathbf{X}=[\mathbf{V}; \mathbf{Q}] \in \mathbb{R}^{B \times (L_c+N) \times d_{\mathrm{enc}}}$. For differentiated treatment in subsequent attention computations, we label text tokens as type 0 and query tokens as type 1.

\subsubsection{Causal-Topological Attention Mask}
Temporal causal reasoning requires simultaneously handling two types of interactions with distinct natures: bidirectional semantic understanding among text tokens and strict causal ordering among event query tokens. To this end, we design a causal-topological attention mask $\mathbf{M}$.

Let the total sequence length be $L=L_c+N$, where positions $1$ to $L_c$ correspond to text tokens and positions $L_c+1$ correspond to query tokens. The mask matrix $\mathbf{M} \in \mathbb{R}^{L \times L}$ is dynamically constructed before the attention computation of each layer. Its element assignment rule is given in Equation (\ref{eq:6}).

\begin{equation}
\mathbf{M}_{i,j} =
\begin{cases} 
0, & \text{if } i, j \leq L_c \text{ (global bidirectional: semantic coherence)} \\
0, & \text{if } i > L_c, j \leq L_c \text{ (queries attend to all text tokens)} \\
0, & \text{if } i, j > L_c \land j \leq i \text{ (global causal)} \\
-\infty, & \text{otherwise}
\end{cases}
\label{eq:6}
\end{equation}

The mask $\mathbf{M}$ defines a mixed attention space. Among text tokens, where $i, j \leq L_c$, all text tokens are visible to each other, achieving global bidirectional attention and ensuring complete contextual understanding of the textual facts. For query tokens attending to text tokens, where $i > L_c, j \leq L_c$, all query tokens can attend to all text tokens, thereby enabling retrieval of event-related information from the full text. Among query tokens, where $i, j > L_c \land j \leq i$, the condition $j \leq i$ ensures that the overall direction of event queries follows a causal order, meaning future queries cannot inform past ones.

\subsubsection{Time-Aware Positional Bias}

To further strengthen the encoder's perception of the physical time dimension, we introduce an explicit time bias matrix $\mathbf{B}_{\mathrm{time}}$. For the query token embeddings $\mathbf{Q}$ and text token embeddings $\mathbf{V}$, we apply learnable linear projections $\mathrm{Proj}_\mathbf{Q}$ and $\mathrm{Proj}_\mathbf{V}$, respectively, and then compute their inner product to obtain the bias matrix $\mathbf{B}_{\mathrm{time}}$, as shown in Equation (\ref{eq:7}).

\begin{equation}
\mathbf{B}_{\mathrm{time}} = \mathrm{Proj}_\mathbf{Q}(\mathbf{Q}) \cdot \mathrm{Proj}_\mathbf{V}(\mathbf{V})^\top
\label{eq:7}
\end{equation}

$\mathbf{B}_{\mathrm{time}}$ is added to the region where $i > L_c, j \leq L_c$ in Equation (\ref{eq:6}), enabling query tokens to adjust their attention weights based on temporal logic when aggregating textual information.

\subsubsection{Auxiliary Timestamp Prediction}

To enable the latent space representation to encode physical temporal positions explicitly, we attach a lightweight timestamp regression head to the encoder. This head consists of two fully connected layers with GELU activation and maps the final hidden state of each query token to a one-dimensional normalized timestamp prediction. The regression head is supervised only during training via mean squared error (MSE) loss, injecting temporal awareness into the latent variables.

\subsubsection{Computation Flow}

The overall computation flow of the encoder can be summarized as follows. The text embeddings $\mathbf{V}$ and event anchor embeddings $\mathbf{Q}$ are concatenated and fed into the backbone network. Before each attention layer computation, the model dynamically constructs the causal topological attention mask $\mathbf{M}$ based on the token type identifiers and adds the time bias matrix $\mathbf{B}_{\mathrm{time}}$. After the network output, the hidden states corresponding to all query tokens are extracted to obtain the causal latent variable sequence $\mathbf{Z}_{\mathrm{causal}} \in \mathbb{R}^{B \times N \times d_{\mathrm{enc}}}$.

\subsection{Reasoning Decoder}

The reasoning decoder is built on the Qwen-3-8B model and aligns the encoder's causal latent variables with user query instructions while performing autoregressive text generation. Since the latent space dimensions and semantic distributions differ between the encoder and decoder, a nonlinear projector $\mathcal{P}$ is placed between them for dimension alignment and representation adaptation. The projector consists of two fully connected layers with GELU activation. To simultaneously leverage the fine-grained semantics of the original factual information and the topological priors in the causal graph structure during reasoning, the decoder adopts a fact-logic bypass fusion strategy. The input embedding sequence of the decoder is formed by concatenating three parts, as shown in Equation (\ref{eq:8}).

\begin{equation}
\mathbf{E}_{\mathrm{input}} = [\mathbf{V}_{\mathrm{text}}; \mathbf{Z}_{\mathrm{aligned}}; \mathbf{E}_{\mathrm{prompt}}]
\label{eq:8}
\end{equation}

where, $\mathbf{V}_{\mathrm{text}}$ is the token embeddings of the original text chunks obtained through the decoder's word embedding layer, which constitutes the factual track. $\mathbf{Z}_{\mathrm{aligned}}$ is the aligned causal latent variables from the encoder output via the projector, which constitutes the logical track. $\mathbf{E}_{\mathrm{prompt}}$ is the embedding representation of the user query instruction.

The corresponding attention mask $\mathbf{A}_{\mathrm{full}}$ is constructed in a block-diagonal form. Within $\mathbf{V}_{\mathrm{text}}$, a standard causal mask is maintained. The $\mathbf{Z}_{\mathrm{aligned}}$ part is globally visible to all positions. The $\mathbf{E}_{\mathrm{prompt}}$ part follows the standard autoregressive mask.

\subsection{Native Counterfactual Intervention}

TCR-Former supports executing counterfactual interventions at the attention mechanism level, without re-encoding the input text or invoking an external graph manipulation engine. The intervention operation is implicitly triggered by counterfactual assumptions in the user query prompt $P$. When the user prompt contains counterfactual statements such as "if an event had not occurred," the system matches them with event anchors in the encoder input. It resolves the query token index $k_\psi$ corresponding to the target event.

After obtaining the index $k_\psi$, the encoder performs topological pruning on the mask matrix $\mathbf{M}$ when constructing the causal-topological attention mask. The element-wise modification rule for the intervention mask $\mathbf{M}^\psi$ is given in Equation (\ref{eq:9}).

\begin{equation}
\mathbf{M}_{i,j}^\psi = \mathbf{M}_{i,j} \cdot \mathbb{I}(i \neq k_\psi) \cdot \mathbb{I}(j \neq k_\psi)
\label{eq:9}
\end{equation}

That is, all incoming and outgoing attention connections of the target query token are cut off by setting the mask values at its corresponding positions to $-\infty$. Through this operation, the causal latent variables $\mathbf{Z}_{\mathrm{causal}}$ output by the encoder implicitly reflect the topological structural changes after the causal chain is broken. In the decoding stage, $\mathbf{Z}_{\mathrm{aligned}}$ carries this intervention information into the autoregressive generation process, and the model reasons about the outcome under the counterfactual scenario based on the latent representations of the remaining causal paths.

\subsection{Forward Computation Flow}

The overall forward computation flow of TCR-Former is presented in Algorithm \ref{alg:2} (Appendix \ref{app: TCR-Agent and TCR-Former Algorithm}). The input consists of the text chunk sequence $C$ and the user query prompt $P$. First, the encoder generates text token embeddings $\mathbf{V}$ and event anchor embeddings $\mathbf{Q}$ with time span biases, concatenates them, computes the time bias matrix, and constructs the causal-topological attention mask. If $P$ contains a counterfactual assumption, the intervention index is resolved, and topological pruning is applied to the mask. The encoder outputs hidden state sequences, from which the causal latent variables $\mathbf{Z}_{\mathrm{causal}}$ corresponding to the query tokens are extracted, and timestamps $\hat{T}$ are predicted via the regression head. Subsequently, $\mathbf{Z}_{\mathrm{causal}}$ is aligned to the decoder space by the projector, yielding $\mathbf{Z}_{\mathrm{aligned}}$. The decoder concatenates the original text embeddings $\mathbf{V}_{\mathrm{text}}$, the aligned latent variables $\mathbf{Z}_{\mathrm{aligned}}$, and the prompt embedding $\mathbf{E}_{\mathrm{prompt}}$, constructs the full attention mask, performs autoregressive decoding, and finally outputs the generated response $Y$.

\section{TempoCausal-Bench}
Existing mainstream temporal benchmarks mostly rely on explicit timestamps to anchor questions, allowing models to answer easily through shallow matching and making it difficult to reach deep temporal causal reasoning. However, complex decision-making and long-range narratives in real-world scenarios precisely involve characteristics such as implicit temporal boundary delineation, concurrent state evolution of multiple entities, and long-range causal lags. To systematically evaluate the topological causal reasoning capabilities of large language models in these scenarios, we build the TempoCausal-Bench by upgrading linear temporal question answering into a structured temporal causal reasoning task through a five-dimensional reconstruction paradigm, based on the difficult samples from TimeQA-Hard \cite{R8}. The detailed construction process of the TempoCausal-Bench is presented in Appendix \ref{app:TempoCausal-Bench}.

\section{Experiments}
\subsection{Experimental Setup}
\textbf{Metrics.} The evaluation metrics adopted are: \textbf{(1) Exact Match (EM):} requires string-identical matching with the standard answer. \textbf{(2) F1-Score:} computing based on token-level overlap to mitigate the strict matching bias for long answers. \textbf{(3) LLM-Acc:} employing DeepSeek-V3.2 \cite{R51} and Gemini-3.1-flash \cite{R52} as judges to perform semantic-level correctness assessment, the detailed evaluation procedures are provided in Appendix \ref{app:LLM-Acc}.

\textbf{Baseline.} We select representative closed-source and open-source models for evaluation, including GPT-5.4 \cite{R53}, DeepSeek-V3.2 \cite{R51}, Qwen-3-8B \cite{R54}, and Qwen-2.5-7B \cite{R55}. In addition, we include methods designed to enhance the temporal question-answering capability of LLMs for comparison, including TG-LLM \cite{R4}, Time-CoT \cite{R56}, and TISER \cite{R57}. For all experiments, these methods are used in their CoT configurations strictly following the prompts publicly disclosed in their papers, and no model training is conducted. Specific introductions to each method are given in Appendix \ref{app:Baseline Methods}. In the TCR-Former experiments, when Qwen-3-8B serves as the decoder, the encoder adopts Qwen-3-0.6B, and when Qwen-2.5-7B serves as the decoder, the encoder adopts Qwen-2.5-0.5B. In the TCR-Agent+LoRA experiment, the TCR-Agent framework is employed, and the model is fine-tuned on question-answer pairs generated by TCR-Agent (GPT-5.4) on the TempoCausal-Bench training set.

\subsection{Main Results}

\begin{wraptable}[25]{r}{0.55\textwidth}
\vspace{-1.5cm}
\centering
\tiny
\definecolor{customBlue}{rgb}{0.855, 0.89, 0.957}
\setlength{\aboverulesep}{0pt}
\setlength{\belowrulesep}{0pt}
\renewcommand{\arraystretch}{1.25}
\caption{TempoCausal benchmark experimental results. The \textbf{Bold} numbers indicate the best performance among all models.}
\label{tab:1}
\vspace{0.35cm}   
\begin{tabular}{l c c c c}
\toprule
Method & Algorithm & EM ($\uparrow$) & F1 ($\uparrow$) & LLM-Acc ($\uparrow$) \\
\midrule
\multicolumn{5}{c}{Train-Free} \\
\midrule
\multirow{5}{*}{GPT-5.4} & Standard & 28.8 & 44.4 & 51.3 \\
 & TG-LLM & 32.0 & 49.3 & 57.1 \\
 & Time-CoT & 30.0 & 47.8 & 54.5 \\
 & TISER & 37.6 & 52.1 & 59.2 \\
 & TCR-Agent (Ours) & 46.9 & 65.6 & 70.4 \\
\midrule
\multirow{5}{*}{Deepseek-V3.2} & Standard & 25.7 & 42.1 & 47.4 \\
 & TG-LLM & 30.3 & 46.1 & 50.4 \\
 & Time-CoT & 31.2 & 46.4 & 50.6 \\
 & TISER & 35.3 & 47.7 & 54.6 \\
 & TCR-Agent (Ours) & 46.4 & 61.4 & 68.2 \\
\midrule
\multirow{5}{*}{Qwen-3-8B} & Standard & 17.7 & 34.3 & 42.5 \\
 & TG-LLM & 21.1 & 37.9 & 46.8 \\
 & Time-CoT & 23.0 & 39.6 & 45.4 \\
 & TISER & 25.2 & 39.1 & 49.2 \\
 & TCR-Agent (Ours) & 39.9 & 51.7 & 60.1 \\
\midrule
\multirow{5}{*}{Qwen-2.5-7B} & Standard & 11.9 & 24.5 & 35.2 \\
 & TG-LLM & 14.6 & 28.5 & 39.6 \\
 & Time-CoT & 13.5 & 28.8 & 37.5 \\
 & TISER & 17.6 & 31.7 & 41.9 \\
 & TCR-Agent (Ours) & 27.8 & 42.1 & 54.8 \\
\midrule
\multicolumn{5}{c}{Train-Based} \\
\midrule
 & TCR-Agent + LoRA & 49.1 & 60.4 & 67.4 \\
\rowcolor{customBlue} \cellcolor{white} \multirow{-2}{*}{Qwen-3-8B} & TCR-Former (Ours) & \textbf{55.9} & \textbf{68.9} & \textbf{75.7} \\
 & TCR-Agent + LoRA & 38.4 & 53.0 & 60.5 \\
\rowcolor{customBlue} \cellcolor{white} \multirow{-2}{*}{Qwen-2.5-7B} & TCR-Former (Ours) & 45.8 & 59.3 & 67.8 \\
\bottomrule
\end{tabular}%
\end{wraptable}

\begin{figure}[t]
\centering
\includegraphics[width=14cm]{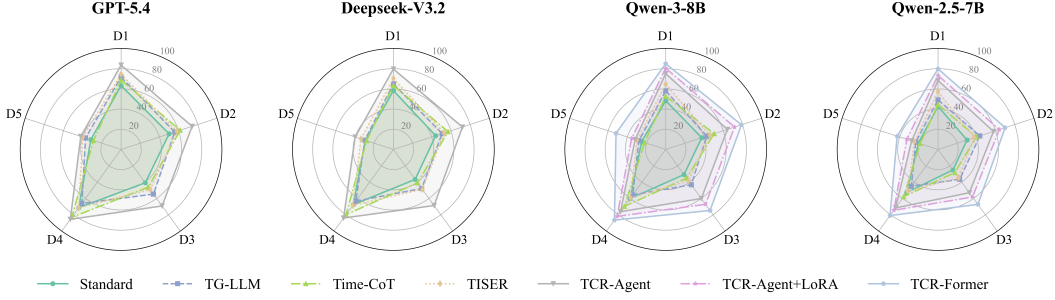}
\caption{Experimental results on each dimension of the TempoCausal-Bench (LLM-Acc).}
\label{fig:2}
\end{figure}

We conduct large-scale comparative experiments on the TempoCausal-Bench. All experiments are independently run three times under identical settings, and the average results are reported to eliminate stochastic fluctuations. Table \ref{tab:1} and Fig. \ref{fig:2} present the performance of each method on the TempoCausal-Bench. The results indicate that complex temporal causal reasoning poses a severe challenge to existing LLMs. Methods such as TG-LLM, Time-CoT, and TISER yield only limited improvements, suggesting that external prompt engineering cannot systematically resolve deep topological reasoning problems. In contrast, TCR-Agent consistently outperforms training-free baselines across all backbone models. When built on GPT-5.4, it attains an LLM-Acc of 70.4\%, an 11.2\% improvement over the best baseline TISER (LLM-Acc = 59.2\%). Furthermore, TCR-Former internalizes structural awareness into model computation, achieving a secondary leap in performance. On Qwen-3-8B, its LLM-Acc, EM, and F1 reach 75.7\%, 55.9\%, and 68.9\%, respectively, comprehensively outperforming all compared methods. To quantify the net gain from architectural internalization, we introduce a fine-tuned TCR-Agent + LoRA (Qwen-3-8B) baseline, which achieves an LLM-Acc of 67.4\%; TCR-Former further improves upon this by 8.3\%. These experimental results demonstrate that designs such as the causal-topological attention mask and time-span biases directly encode topological reasoning priors into the forward computation, effectively bridging the semantic gap in cross-modal and neuro-symbolic transformations.

\textbf{Out-of-Domain Experiment.} To evaluate the models' generalization capability, we conduct tests on the widely used temporal reasoning benchmarks TempReason (L3) \cite{R58} and TGQA \cite{R4}. The experimental results show that our proposed methods surpass existing SOTA methods, where TCR-Former achieves an EM of 95.4\% on TempReason (L3) and an EM of 91.9\% on TGQA. The complete experimental results are presented in Appendix \ref{app:Out-of-Domain Experiment}.

\begin{wraptable}[12]{r}{0.6\textwidth}
    \vspace{-1.5cm}          
    \raggedleft 
    \tiny
    \definecolor{customBlue}{rgb}{0.855, 0.89, 0.957}
    \setlength{\aboverulesep}{0pt}
    \setlength{\belowrulesep}{0pt}
    \renewcommand{\arraystretch}{1.25}
    \setlength{\tabcolsep}{3pt}
    \caption{Ablation experiment results.}
    \label{tab:2}
    \vspace{0.35cm}   
    \begin{tabular}{l c c c c}
        \toprule
        Component & Architecture Setting & EM ($\uparrow$) & F1 ($\uparrow$) & LLM-Acc ($\uparrow$) \\
        \midrule
        \multirow{6}{*}{\begin{tabular}{@{}l@{}}TCR-Agent \\ Base: GPT-5.4\end{tabular}} 
         & Full TCR-Agent & 46.9 & 65.6 & 70.4 \\
         & w/o Scratchpad & 43.2 & 60.1 & 66.8 \\
         & w/o DAG Instantiation & 40.1 & 55.3 & 63.1 \\
         & w/o Counterfactual Intervention & 44.0 & 62.7 & 67.5 \\
         & w/o Dual-Track Reasoning & 41.3 & 58.6 & 65.2 \\
         & Standard & 28.8 & 44.4 & 51.3 \\
        \midrule
        \multirow{7}{*}{\begin{tabular}{@{}l@{}}TCR-Former \\ Base: Qwen-3-8B\end{tabular}} 
         & Full TCR-Former & 55.9 & 68.9 & 75.7 \\
         & w/o Time Span Bias & 52.4 & 65.2 & 72.3 \\
         & w/o Fact-Logic Bypass Fusion & 50.7 & 64.1 & 71.4 \\
         & w/o Causal-Topological Attention Mask & 48.2 & 60.4 & 67.0 \\
         & Only Decoder & 38.5 & 50.2 & 58.6 \\
         & Only Encoder + Decoder & 34.2 & 47.7 & 55.4 \\
         & Standard & 17.7 & 34.3 & 42.5 \\
        \bottomrule
    \end{tabular}
\end{wraptable}

\subsection{Ablation Study}
To quantify the contributions of each core component, we conduct ablation studies on the TempoCausal Benchmark. The results are presented in Table \ref{tab:2}, and the detailed settings of the ablation study are provided in Appendix \ref{app:Ablation Study Setup}. The experimental results show that, in the TCR-Agent architecture, removing DAG Instantiation leads to an LLM-Acc drop of 7.3\%, the largest decline, indicating that converting text into a rigorous topological structure is a crucial prerequisite for overcoming temporal blind spots. In the TCR-Former architecture, removing the Causal-Topological Attention Mask causes an 8.7\% drop in LLM-Acc, the largest decline among component ablations, proving that severing the bidirectional context of concurrent events severely damages the native perception of causal logic. Notably, Only Decoder achieves 58.6\% LLM-Acc, substantially outperforming the 42.5\% obtained by the Standard baseline. Whereas Only Encoder + Decoder falls to 55.4\%, confirming that the superior performance of Full TCR-Former stems from architectural design rather than training data or encoder parameters. The detailed analysis of the Causal-Topological Attention Mask is presented in Appendix \ref{app:Causal-Topological Attention Mask Analysis}.

\begin{wrapfigure}[28]{r}{6cm}
\vspace{-2.2cm} 
\centering
\includegraphics[width=5.8cm]{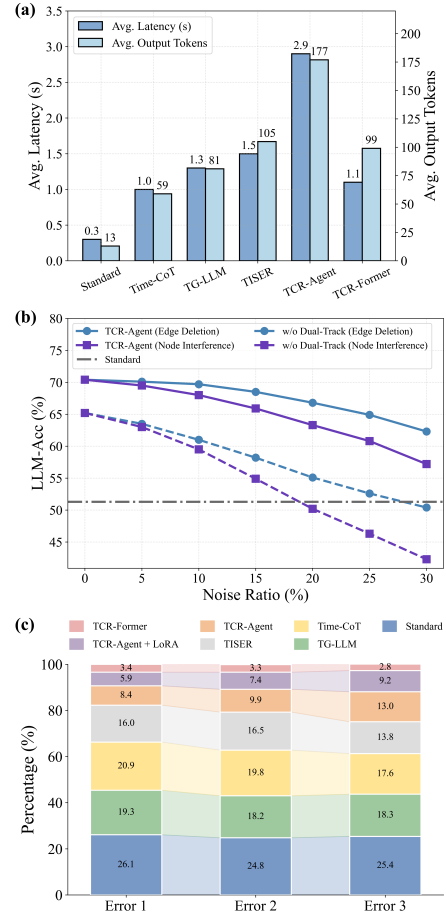}
\vspace{-0.1cm}
\caption{Indicator visualization. (a): Comparison of reasoning efficiency. (b): Robustness analysis of causal graphs. (c): Error Percentage.}
\label{fig:3}
\end{wrapfigure}

\section{Analysis}

\subsection{Reasoning Efficiency Analysis}
To evaluate the computational efficiency of each method in practical deployment, we test the inference overhead of various methods on the TempoCausal-Bench using Qwen-3-8B. The results are shown in Fig. \ref{fig:3}(a). TCR-Agent incurs the highest end-to-end latency and token consumption, as it requires LLM calls for graph extraction and dual-track reasoning. In contrast, TCR-Former internalizes causal-topological awareness and time biases entirely into the model’s forward computation, reducing inference latency by 62\% and output token count by approximately 44\% compared with TCR-Agent. Its efficiency is comparable to that of TISER. A complete experimental analysis is provided in Appendix \ref{app:Reasoning Efficiency Analysis}.

\subsection{Causal Graph Robustness Analysis}
To evaluate the impact of causal graph extraction errors on downstream reasoning, we randomly select 500 samples from TempoCausal and conduct noise injection experiments based on TCR-Agent (GPT-5.4) and its ablation variant (w/o Dual-Track Reasoning). Specifically, we design two perturbation paradigms:

\textbf{(1) Edge Deletion:} randomly removing directed edges to simulate dependency omission. 

\textbf{(2) Node Interference:} replacing node text with irrelevant, distracting text to simulate entity semantic distortion. 

The experimental results are shown in Fig. \ref{fig:3}(b). The results indicate that interference affecting nodes has a greater impact than interference affecting edges, and that the Dual-Track Reasoning mechanism of TCR-Agent helps improve the system’s robustness. A complete experimental analysis is provided in Appendix \ref{app:Causal Graph Robustness Analysis}.

\subsection{Error Attribution Analysis}
\label{Error Attribution Analysis}
To systematically analyze the failure mechanisms of models in complex temporal tasks, we categorize error attributions into four primary error patterns: Temporal Boundary Misjudgment (Error 1), Causal Chain Breakage (Error 2), Rule Inference Mismatch (Error 3), and Other Errors (Error 4). Detailed explanations of the various errors and a detailed analysis of this section are provided in Appendix \ref{app:Error Attribution Analysis}.

Based on Qwen-3-8B, we randomly sample 200 instances per method and perform error attribution. Fig. \ref{fig:3}(c)  presents the error distribution for each method. Upon deeper analysis of the experimental results, we observe a noteworthy \textbf{structure-reasoning disconnection phenomenon}: external graph structures and CoT reasoning can make the model’s reasoning structure more precise, but they fail to fundamentally address the model’s deficiency in deductive capability. As shown in Fig. \ref{fig:3}(c), compared with existing methods, TCR-Agent drastically reduces the error proportions of temporal boundary misjudgment (8.4\%) and causal chain breakage (9.9\%). However, rule inference mismatch remains as high as 13.0\%, nearly on par with TISER (13.8\%). This indicates that combining external graph structures with CoT for reasoning still struggles to handle error types such as Rule Inference Mismatch, which demand deeper deductive capabilities. In contrast, TCR-Former keeps the proportions of all three error types below 3.4\%, with rule inference mismatch at only 2.8\%, indicating that internalizing temporal causal priors into the Transformer attention space can effectively alleviate the structure-reasoning disconnection problem.

\section{Related Work}
Existing research on temporal causal reasoning primarily centers on two main paradigms. In neuro-symbolic reasoning, approaches leveraging graph structures and symbolic working memory \cite{R37,R41,R45} effectively enhance multi-hop compositional stability, yet they predominantly target static commonsense domains, lacking dedicated mechanisms for dynamic temporal evolution and counterfactual deduction. In temporal reasoning for large language models, existing approaches, including timestamp injection, temporal graphs, and chain-of-thought prompting \cite{R5,R8,R11,R26,RN3}, predominantly remain confined to superficial chronological ordering, failing to internalize temporal span biases and causal topological constraints into the foundational self-attention computation of the model. Furthermore, at the underlying computational architecture level, this work draws critical inspiration from layout rearrangement studies in visual document understanding. This field has progressed from early spatial coordinate modeling \cite{R18} to recent paradigms that directly encode reading order topology into forward computation via attention biases, such as ROAP \cite{R20} and DeepSeek-OCR 2 \cite{R21}. This motivates us to design the Causal-Topological Attention Mask and temporal span biases, transferring topology awareness into the self-attention layer to achieve end-to-end temporal causal modeling. The complete related work is detailed in Appendix \ref{app:Complete Related Work}.

\section{Conclusion}
We propose TCR-Agent and TCR-Former, two systematic approaches that, respectively, follow the training-free neuro-symbolic framework and the natively internalized causal architecture to tackle temporal causal reasoning in complex narratives. On the TempoCausal-Bench, TCR-Agent consistently surpasses existing training-free baselines across multiple models. Moreover, TCR-Former achieves 75.7\% accuracy with approximately 8.6B parameters, outperforming the previous best method by 16.5\%, and breaking through the structure-reasoning disconnection bottleneck under external symbolic constraints. Ablation studies validate the effectiveness of each component in TCR-Agent and TCR-Former, providing an efficient and reliable reasoning paradigm for real-world decision-making scenarios such as hypothesis deduction with strong temporal dependencies and long-range lag prediction.

\section*{Acknowledgements}
This work was supported by the Development Agreement between Saudi Aramco Technologies Company and South China University of Technology (SCUT) (SATC 2025-024) and by the Introduced Innovative R\&D Team of Guangdong (2023ZT10L145), and it used the facilities of Pazhou Laboratory. All opinions expressed in this paper are the authors’ and do not necessarily reflect the policies and views of the sponsors.

\bibliographystyle{unsrtnat}
\bibliography{references}

\newpage
\appendix
\section{Complete Related Work}
\label{app:Complete Related Work}
\textbf{Structural Rearrangement and Topology Modeling in Document Visual Understanding.} In document visual understanding, the core challenge lies in reconstructing discrete and disordered visual elements on a two-dimensional layout into a one-dimensional sequence conforming to a logical reading order \cite{R61,R62,R63}. LayoutLM \cite{R18} introduces spatial coordinate embeddings to provide preliminary modeling of document layout, yet subsequent studies indicate that relying solely on geometric position information is insufficient to capture complex layout structures \cite{R29,R32}. To address this, researchers have introduced hierarchical structures \cite{R30}, markup languages \cite{R31}, and other techniques for structured modeling, and have gradually established reading order as a key variable linking spatial layout and semantic logic \cite{R19,R33,R34,R35,R36}. Recent methods such as ROAP \cite{R20} and DeepSeek-OCR 2 \cite{R21} further encode topology priors related to reading order directly into the attention computation of Transformers via structure-aware attention biases and query mechanisms, achieving end-to-end rearrangement of unordered visual blocks into an ordered logical sequence. These studies collectively demonstrate that explicitly encoding latent structures into attention \cite{R21} is more effective than simple spatial modeling, inspiring us to transfer topology-aware attention design to temporal causal reasoning.

\textbf{Graph Structures and Neuro-Symbolic Methods for Complex Reasoning.} To alleviate the lost-in-the-middle problem in long-context reasoning \cite{R8}, researchers have begun exploring graph structures and neuro-symbolic methods. In retrieval-augmented generation, path planning and subgraph retrieval based on knowledge graphs have effectively enhanced multi-hop reasoning capabilities \cite{R37,R38,R39,R40,R41,RN1,RN2}. In broader complex reasoning tasks, explicitly encoding entity relations via graph structures \cite{R42,R43,R44} or introducing symbolic working memory \cite{R45,R46} has improved the stability of models in handling cross-task compositional rules. However, the above methods are primarily oriented towards non-temporal scenarios, such as general reasoning and commonsense question answering, and lack targeted designs for complex temporal dependencies and counterfactual reasoning mechanisms. In contrast, the TCR-Agent framework proposed in this paper explicitly models temporal causal topology through an external causal graph and interventional topology pruning.

\textbf{Temporal Perception and Causal Reasoning in LLMs.} Standard retrieval-augmented generation (RAG) systems typically treat temporal information as shallow textual features, struggling to handle the dynamic evolution of events and long-range dependencies \cite{R1,R2,R7,R8}. Early methods relying on timestamp injection or multi-turn prompting \cite{R7,R8} remain at the level of superficial temporal ordering and cannot support deep causal modeling \cite{R5,R8,R23}, a limitation also revealed by multiple benchmarks such as TRAM \cite{R24}, ETRQA \cite{R25}, and CausalBench \cite{R26}. Although subsequent studies have introduced strategies such as temporal graphs \cite{R11,RN3}, multi-hop data \cite{R27}, and instruction tuning \cite{R28} to enhance structured temporal representations, these methods still essentially depend on external prompts or intermediate structures and do not internalize temporal biases and causal constraints into the core attention mechanism of the model. In contrast to these works, TCR-Former introduces temporal span biases and a causal-topological attention mask into the attention layers, thereby fundamentally reconstructing a nonlinear causal attention mechanism with topology-awareness.

\section{Limitation and Future Work}
Although the proposed TCR-Agent and TCR-Former demonstrate significant performance advantages in complex temporal causal reasoning tasks and effectively alleviate the core "structure-reasoning disconnection" dilemma in existing methods, this study still has several limitations that warrant further exploration and refinement in future work.

\textbf{Limitation.} In terms of computational efficiency and training paradigms, TCR-Agent relies on multiple calls to large language models and cascaded scheduling of an external symbolic engine, resulting in high end-to-end inference latency that makes it difficult to meet high-throughput, low-latency real-time decision-making requirements. While TCR-Former reduces inference overhead by approximately 62\% through architectural internalization, its three-stage training strategy is relatively complex. It has currently been validated mainly on Qwen-series backbones, leaving its cross-architecture generalization yet to be further verified. Second, regarding the causal representation construction mechanism, the causal graph instantiation in this paper heavily relies on large language models' text-extraction capabilities, resulting in an essentially "narrative-driven" topological reconstruction. In long texts with implicit information, ambiguous references, or conflicting expressions, hallucinations or structural biases during extraction may propagate downstream through the DAG, thereby affecting the reliability of counterfactual intervention. Third, regarding the benchmark's multimodal coverage, the proposed TempoCausal-Bench currently focuses solely on the plain-text modality. However, numerous temporal causal reasoning tasks in real-world scenarios involve multimodal signals such as text, charts, and numerical sequences. The existing benchmark cannot measure the model's ability to fuse cross-modal information and perform collaborative causal inference, limiting the extrapolation and validation of architectures such as TCR-Former for practical multimodal applications.

\textbf{Future Work.} To address the above limitations, future work will proceed along the following directions. To improve efficiency, we will explore end-to-end joint optimization or representation distillation strategies to simplify the training pipeline and introduce sparsification or linear-complexity variants to reduce the computational overhead of dynamic masking. Meanwhile, we will attempt to establish formal connections between the Causal-Topological Attention mechanism and structural causal models, thereby providing theoretical bounds on the identifiability and error propagation in counterfactual deduction. We will further investigate causal discovery and uncertainty modeling, extending beyond the pure text extraction paradigm. By integrating symbolic causal discovery algorithms or incorporating Bayesian graph structure learning, we will probabilistically model causal edge weights and intervention paths, thereby quantifying the structural uncertainty during topology construction and enhancing the robustness of counterfactual truncation. To expand the multimodal benchmark, we plan to extend the evaluation dimensions of the TempoCausal-Bench to include multimodal scenarios, constructing a temporal causal reasoning evaluation dataset that includes modalities such as images, structured tables, and time-series plots. By designing tasks such as cross-modal event alignment, multi-source information conflict resolution, and chart-based trend reasoning, we will systematically assess the model's capability in causal topology reconstruction and counterfactual deduction in complex multimodal environments, providing a standardized testbed for developing a unified multimodal temporal causal reasoning backbone.

\section{Experiments and Analysis}
\subsection{Out-of-Domain Experiment}
\label{app:Out-of-Domain Experiment}
To evaluate the models' generalization capability, we conduct extended tests on two widely used temporal reasoning benchmarks, TempReason (L3) \cite{R58} and TGQA \cite{R4}. TCR-Agent uses GPT-5.4 as the base model, and TCR-Former employs Qwen-3-8B as the reasoning decoder. The detailed experimental results are presented in Table \ref{tab:3}.

On the TempReason (L3) benchmark, the experimental results indicate that both methods proposed in this paper surpass existing strong baselines. As a classic temporal question answering evaluation benchmark, TempReason (L3) primarily relies on explicit temporal anchors, focusing on assessing chronological ordering and temporal commonsense computation, while lacking deep causal topology and counterfactual deduction. The training-free TCR-Agent achieves an EM of 93.2\% and an F1 score of 95.6\%, outperforming the previous best baseline, TISER (EM = 91.5\%, F1 = 94.9\%). Building upon this, the end-to-end architecture TCR-Former further attains an EM of 95.4\% and an F1 score of 97.1\%, achieving optimal performance across all metrics.

On the TGQA benchmark, which places greater emphasis on complex multi-entity temporal relation determination, our proposed methods likewise demonstrate significant performance advantages. TGQA requires models to accurately resolve overlapping events, lifespans, and relative durations within long-context multi-entity narratives. Experiments show that the EM and F1 of TCR-Agent reach 91.4\% and 96.6\%, representing improvements of 6.9\% and 2.4\% over TISER (EM = 84.5\%, F1 = 94.2\%), respectively. Furthermore, TCR-Former refreshes the state-of-the-art performance with an EM of 91.9\% and an F1 of 97.5\%. This demonstrates that the proposed architectures possess exceptional generalizability beyond complex causal deduction, confirming that effectively injecting causal topology and temporal biases into the model's reasoning mechanism fundamentally reinforces the language model's temporal causal perception and reasoning capabilities.

\begin{table}[htbp]
    \centering
    \small
    \definecolor{customBlue}{rgb}{0.855, 0.89, 0.957}
    \setlength{\aboverulesep}{0pt}
    \setlength{\belowrulesep}{0pt}
    \renewcommand{\arraystretch}{1.38}
    \setlength{\tabcolsep}{8pt}
    \caption{Out-of-domain experimental results. Results with $^\dagger$ are reported in the original papers \cite{R57}.}
    \label{tab:3}
    \begin{tabular}{lcccc}
        \toprule
        \multirow{2}{*}{Method} & \multicolumn{2}{c}{TempReason (L3)} & \multicolumn{2}{c}{TGQA} \\
        \cmidrule(lr){2-3} \cmidrule(lr){4-5}
                                & EM   & F1   & EM   & F1   \\
        \midrule
        T5-large\cite{R59}$^\dagger$     & 28.8 & 46.8 & 54.8 & 71.3 \\
        Temp-T5\cite{R58}$^\dagger$      & 26.1 & 43.0 & 64.0 & 77.8 \\
        REMEMO-large\cite{R60}$^\dagger$ & 33.4 & 49.3 & 46.1 & 66.0 \\
        TG-LLM\cite{R4}$^\dagger$        & 35.6 & 46.9 & 79.7 & 85.0 \\
        TISER\cite{R57}$^\dagger$        & 91.5 & 94.9 & 84.5 & 94.2 \\
        \midrule
        TCR-Agent                        & 93.2 & 95.6 & 91.4 & 96.6 \\
        TCR-Former                       & 95.4 & 97.1 & 91.9 & 97.5 \\
        \bottomrule
    \end{tabular}
\end{table}

\subsection{Reasoning Efficiency Analysis}
\label{app:Reasoning Efficiency Analysis}
To systematically evaluate the computational efficiency of each method in practical deployment, we test the inference overhead of various methods on the TempoCausal-Bench using Qwen-3-8B, with a batch size of 1 to simulate real-world deployment scenarios, running on 2 × NVIDIA A100-40G GPUs. The results are shown in Fig. \ref{fig:3}(a). The standard prompt exhibits the lowest latency and output token consumption. Time-CoT, TG-LLM, and TISER introduce additional steps such as timeline explicitation, structured generation, and multi-stage self-reflection, leading to progressively increasing token outputs and correspondingly higher latency. TCR-Agent incurs the highest end-to-end latency and token consumption, as it requires LLM calls for graph extraction and dual-track reasoning, along with topological sorting and intervention pruning executed by an external symbolic engine. In contrast, TCR-Former internalizes causal-topological awareness and time biases entirely into the model’s forward computation, reducing inference latency by 62\% and output token count by approximately 44\% compared with TCR-Agent. Its efficiency is comparable to that of TISER, effectively overcoming the efficiency bottleneck of training-free neuro-symbolic frameworks in large-scale deployment.

\subsection{Causal Graph Robustness Analysis}
\label{app:Causal Graph Robustness Analysis}
To evaluate the impact of causal graph extraction errors on downstream reasoning, we randomly select 500 samples from TempoCausal and conduct noise injection experiments based on TCR-Agent (GPT-5.4) and its ablation variant (w/o Dual-Track Reasoning). Specifically, we design two perturbation paradigms:

\textbf{(1) Edge Deletion:} randomly removing directed edges to simulate dependency omission. 

\textbf{(2) Node Interference:} replacing node text with irrelevant, distracting text to simulate entity semantic distortion.

Fig. \ref{fig:3}(b) illustrates the performance degradation curves under the two types of noise. After removing dual-track reasoning, the system’s sensitivity to noise increases significantly. Taking the 15\% noise ratio as an example, under edge deletion, the accuracy of the complete TCR-Agent decreases from 70.4\% to 68.5\% (a drop of 1.9\%), whereas that of the ablation variant drops sharply from 65.2\% to 58.2\% (a drop of 7.0\%). The contrast is even more pronounced under node interference: at a high noise ratio of 30\%, the complete TCR-Agent still maintains an accuracy of 57.2\%, while the ablation variant plummets to 42.3\%, far below the Standard baseline without a causal graph (51.3\%). These results indicate that the original text pathway in dual-track reasoning provides critical fault tolerance against causal graph noise. Through fact-logic dual-track synergy, TCR-Agent can correct deviations by referencing the original text even when the causal graph is corrupted, thereby achieving significantly enhanced robustness.

\subsection{Error Attribution Analysis}
\label{app:Error Attribution Analysis}
To systematically analyze the failure mechanisms of models in complex temporal tasks, we summarize error attribution into four primary error patterns.

\textbf{(1) Temporal Boundary Misjudgment:} inability to accurately locate the start and end ranges of implicit temporal windows or concurrent states, corresponding to typical error types in dimensions D1 and D2. 

\textbf{(2) Causal Chain Breakage:} omitting critical nodes, reversing chronological order, or incurring intervention pruning failure in multi-hop dependencies, predominantly manifested in dimension D3. 

\textbf{(3) Rule Inference Mismatch:} when faced with hypothetical questions or long-cycle lagged deduction, the model ignores explicit calculation rules and quantitative constraints given in the text, instead relying on superficial word associations or general commonsense to make erroneous inferences, leading to deduction results that severely violate the original textual logic; concentrated in dimensions D4 and D5. 

\textbf{(4) Other Errors:} errors not falling into the above categories.

Based on Qwen-3-8B, we randomly sample 200 instances per method and perform error attribution. Fig. \ref{fig:3}(c) presents the distribution of error types for each method. The Standard prompt and the three prompt-enhanced baselines all exhibit high proportions across the three primary error types. Although TG-LLM and TISER achieve slight improvements by introducing graph structures or reflection mechanisms, the overall gains are limited. Upon deeper analysis of the experimental results, we observe a noteworthy structure-reasoning disconnection phenomenon: external graph structures and CoT reasoning can make the model's reasoning structure more precise, but they fail to address the model's deficiency in deductive capability fundamentally. As shown in Fig. \ref{fig:3}(c), compared with existing methods, TCR-Agent drastically reduces the error proportions of temporal boundary misjudgment (8.4\%) and causal chain breakage (9.9\%). Yet, rule inference mismatch remains as high as 13.0\%, nearly on par with TISER (13.8\%). Even when LoRA fine-tuning is applied on top of TCR-Agent, rule inference mismatch only drops to 9.2\%, significantly higher than the other two error types (5.9\% and 7.4\%), further corroborating the structure-reasoning disconnection phenomenon. This indicates that relying solely on external graph structures for reasoning still struggles to handle error types like Rule Inference Mismatch that demand deeper deductive capabilities. In contrast, TCR-Former keeps the proportions of all three error types below 3.4\%, with rule inference mismatch at only 2.8\%, achieving balanced, substantial convergence across all dimensions. The above results indicate that relying solely on external structural constraints can achieve only static reconstruction of the causal topology, whereas embedding temporal causal priors into the attention flow and representation update mechanisms can simultaneously overcome the dual challenges of precise reasoning structure and insufficient deductive capability.

\subsection{Causal-Topological Attention Mask Analysis}
\label{app:Causal-Topological Attention Mask Analysis}
This section further examines the underlying mechanism of the Causal-Topological Attention Mask to verify whether its performance gains stem from actionable, structured causal topological representations rather than a superficial temporal ordering prior. Building upon the ablation study and error attribution analysis, we conduct a dimension-level ablation analysis and an internal failure mode investigation on the variant without this mask.

Table \ref{tab:c5_1} summarizes the answering accuracy (LLM-Acc) across the five core evaluation dimensions of the TempoCausal-Bench under the ablation of the Causal-Topological Attention Mask. Compared with the Full TCR-Former, removing the causal-topological mask leads to performance drops of 2.9\%, 5.5\%, 9.1\%, 16.2\%, and 9.3\% across dimensions D1 to D5, respectively, exhibiting the most pronounced declines in D4 and D5. If the mask merely provided a temporal ordering prior, its performance advantages would primarily concentrate on shallow tasks that heavily rely on temporal anchor alignment, specifically D1 and D2. However, the performance degradation on D1 and D2 upon mask removal remains relatively moderate, at 2.9\% and 5.5\%, respectively. In contrast, accuracy on the deep causal task D4 drops sharply by 16.2\%, accompanied by substantial degradation in D3 and D5. This pronounced asymmetry in performance degradation indicates that the standard autoregressive mask is sufficient for handling pure event sequencing, whereas the performance advantage of the causal-topological mask primarily arises from sustaining the causal pathways required for counterfactual deduction and supporting structured intervention capabilities.

\begin{table}[htbp]
    \centering
    \small
    \definecolor{customBlue}{rgb}{0.855, 0.89, 0.957}
    \setlength{\aboverulesep}{0pt}
    \setlength{\belowrulesep}{0pt}
    \renewcommand{\arraystretch}{1.25}
    \setlength{\tabcolsep}{8pt}
    \caption{Ablation of Causal-Topological Attention Mask (LLM-Acc). D1: Implicit temporal windows and conditional filtering. D2: Concurrent multi-entity state tracking. D3: Temporal lags and multi-hop dependencies. D4: Counterfactual cascading deduction. D5: Lagged amplification of minute perturbations.}
    \label{tab:c5_1}
    \begin{tabular}{llccccc}
        \toprule
        Component & Architecture Setting & D1 & D2 & D3 & D4 & D5 \\
        \midrule
        \multirow{3}{*}{\shortstack[l]{TCR-Former\\Base: Qwen-3-8B}} 
                  & Full TCR-Former                       & 85.0 & 79.0 & 74.8 & 86.5 & 51.8 \\
                  & w/o Causal-Topological Attention Mask & 82.1 & 73.5 & 65.7 & 70.3 & 42.5 \\
                  & Standard                              & 48.3 & 37.6 & 30.8 & 60.4 & 25.1 \\
        \bottomrule
    \end{tabular}
\end{table}

Regarding internal failure modes, building upon the error attribution analysis in Section \ref{Error Attribution Analysis}, we supplement the error distribution for this ablation variant to dissect the specific role of the Causal-Topological Attention Mask, as shown in \ref{tab:c5_2}. After removing the causal-topological mask, Error 2 and Error 3 increase substantially from 3.1\% and 2.6\% to 11.1\% and 12.3\%, respectively. In contrast, Error 1, which primarily reflects event boundary localization and associative ordering capabilities, increases only modestly from 3.3\% to 6.1\%. The marginal increase in Error 1 indicates that sequential ordering degradation is not the primary factor driving overall performance deterioration. Conversely, the substantial surge in Error 2 and Error 3 corresponds to the severance of causal dependencies and the impairment of deep deductive capabilities. This demonstrates that the intrinsic function of the causal-topological mask is to preserve directed causal connections among events within the attention space, thereby supporting the model in executing causally consistent deduction under interventional and counterfactual conditions, rather than merely enforcing chronological event constraints.

\begin{table}[htbp]
    \centering
    \small
    \definecolor{customBlue}{rgb}{0.855, 0.89, 0.957}
    \setlength{\aboverulesep}{0pt}
    \setlength{\belowrulesep}{0pt}
    \renewcommand{\arraystretch}{1.3}
    \setlength{\tabcolsep}{10pt}
    \caption{Error proportions under mask ablation (\%). Error 1: Temporal Boundary Misjudgment. Error 2: Causal Chain Breakage. Error 3: Rule Inference Mismatch.}
    \label{tab:c5_2}
    \begin{tabular}{llccc}
        \toprule
        Component & Architecture Setting & Error 1 & Error 2 & Error 3 \\
        \midrule
        \multirow{3}{*}{\shortstack[l]{TCR-Former\\Base: Qwen-3-8B}} 
                  & Full TCR-Former                       & 3.3  & 3.1  & 2.6  \\
                  & w/o Causal-Topological Attention Mask & 6.1  & 11.1 & 12.3 \\
                  & Standard                              & 24.3 & 22.1 & 22.3 \\
        \bottomrule
    \end{tabular}
\end{table}

\section{TCR-Agent and TCR-Former Algorithm}
\label{app: TCR-Agent and TCR-Former Algorithm}

The core algorithm process of TCR-Agent is shown in Algorithm \ref{alg:1}.

\begin{center}
\begin{minipage}{0.65\textwidth} 
\begin{algorithm}[H]
\small
\caption{TCR-Agent Temporal-Causal Reasoning}
\label{alg:1}
\textbf{Input:} Text chunks $C$, Prompt $P$ \\
\textbf{Output:} Final answers $\hat{\mathbf{A}}_q$
\vspace{0.05in}
\hrule
\vspace{0.05in}

\linespread{1.1}\selectfont 

\begin{algorithmic}[1]
    \State $\triangleright$ \textbf{Stage 1: Causal Graph Extraction}
    \State $(S, G_{\mathrm{raw}}) \leftarrow LLM_\theta(C, P)$
    \State $G \leftarrow B(G_{\mathrm{raw}})$
    \State $\triangleright$ \textbf{Stage 2: Counterfactual Intervention}
    \State $\psi \leftarrow P$
    \If{$\psi \neq \emptyset$}
        \State $G' \leftarrow \mathcal{I}(G, \psi)$
    \Else
        \State $G' \leftarrow G$
    \EndIf
    \State $D(G') \leftarrow \mathrm{GraphInterventionEngine.serialize}(G')$
    \State $\triangleright$ \textbf{Stage 3: Dual-Track Reasoning}
    \State $\hat{\mathbf{A}}_q \leftarrow LLM_\varphi(C, D(G'), P)$
    \State \Return $\hat{\mathbf{A}}_q$
\end{algorithmic}
\end{algorithm}
\end{minipage}
\end{center}

The core algorithm process of TCR-Former is shown in Algorithm \ref{alg:2}.

\begin{center}
\begin{minipage}{1.0\textwidth} 
\begin{algorithm}[H] 
\small
\caption{TCR-Former Temporal-Causal Forward Pass}
\label{alg:2}
\textbf{Input:} Text chunks $C$, Prompt $P$ \\
\textbf{Output:} Generated response $Y$, Predicted timestamps $\hat{T}$
\vspace{0.05in}
\hrule
\vspace{0.05in}

\linespread{1.1}\selectfont 

\begin{algorithmic}[1]
    \State $\triangleright$ \textbf{Stage 1: Temporal Causal Encoding}
    \State $\mathbf{V} \leftarrow \mathrm{EmbedText}(C)$; $\mathbf{Q} \leftarrow \mathrm{AnchorEmbedding} + \mathrm{TimeSpanBias}$
    \State $\mathbf{X} \leftarrow \mathrm{Concatenate}(\mathbf{V}, \mathbf{Q})$
    \State $\mathbf{B}_{\mathrm{time}} \leftarrow \mathrm{ComputeTemporalBias}(\mathrm{Proj}_Q(\mathbf{Q}), \mathrm{Proj}_V(\mathbf{V}))$
    \State $\mathbf{M} \leftarrow \mathrm{BuildDynamicCausalMask}(L_c, N)$
    \State $\mathbf{M} \leftarrow \mathrm{InjectTemporalBias}(\mathbf{M}, \mathbf{B}_{\mathrm{time}})$
    \If{$P$ contains counterfactual assumption}
        \State $k_\psi \leftarrow \mathrm{ParseInterventionTarget}(P, \mathrm{event\_list})$
        \State $\mathbf{M} \leftarrow \mathrm{ExecuteIntervention}(\mathbf{M}, k_\psi)$
    \EndIf
    \State $\mathbf{H}_{\mathrm{enc}} \leftarrow \mathrm{ForwardEncoder}(\mathbf{X}, \mathrm{attention\_mask} = \mathbf{M})$ \quad \text{// Causal-Topological Attention}
    \State $\mathbf{Z}_{\mathrm{causal}} \leftarrow \mathrm{ExtractQueryStates}(\mathbf{H}_{\mathrm{enc}}, \mathrm{offset} = L_c)$ \quad \text{// Latent causal sequence}
    \State $\hat{T} \leftarrow \mathrm{RegressTimestamp}(\mathbf{Z}_{\mathrm{causal}})$ \quad \text{// Auxiliary timestamp prediction head}
    \State $\triangleright$ \textbf{Stage 2: Reasoning \& Bypass Fusion}
    \State $\mathbf{Z}_{\mathrm{aligned}} \leftarrow \mathrm{ProjectToDecoderSpace}(\mathbf{Z}_{\mathrm{causal}})$ \quad \text{// Nonlinear alignment}
    \State $\mathbf{V}_{\mathrm{text}} \leftarrow \mathrm{EmbedText}(C)$; $\mathbf{E}_{\mathrm{prompt}} \leftarrow \mathrm{EmbedPrompt}(P)$
    \State $\mathbf{E}_{\mathrm{input}} \leftarrow \mathrm{Concatenate}(\mathbf{V}_{\mathrm{text}}, \mathbf{Z}_{\mathrm{aligned}}, \mathbf{E}_{\mathrm{prompt}})$
    \State $\mathbf{A}_{\mathrm{full}} \leftarrow \mathrm{MergeAttentionMasks}(\mathbf{V}_{\mathrm{text}}, \mathbf{Z}_{\mathrm{aligned}}, \mathbf{E}_{\mathrm{prompt}})$
    \State $Y \leftarrow \mathrm{AutoregressiveDecode}(\mathbf{E}_{\mathrm{input}}, \mathrm{attention\_mask} = \mathbf{A}_{\mathrm{full}})$ \quad \text{// Constrained generation}
    \State \Return $Y, \hat{T}$
\end{algorithmic}
\end{algorithm}
\end{minipage}
\end{center}

\section{TempoCausal-Bench}
\label{app:TempoCausal-Bench}
\subsection{Five-Dimensional Evaluation}
Existing mainstream temporal benchmarks mostly rely on explicit timestamps to anchor questions, allowing models to answer easily through shallow matching and making it difficult to reach deep temporal causal reasoning. We build the TempoCausal-Bench by upgrading linear temporal question answering into a structured temporal causal reasoning task through a five-dimensional reconstruction paradigm, based on the difficult samples from TimeQA-Hard \cite{R8}. The specific dimensions are introduced as follows.

\textbf{(1) Implicit Temporal Windows and Conditional Filtering.} Existing benchmarks often directly provide explicit time ranges, weakening the model's autonomous ability to perceive event boundaries. This dimension aims to test the model's reasoning ability to delineate temporal boundaries through event anchors without explicit dates.

\textbf{(2) Concurrent Multi-Entity State Tracking.} Traditional temporal data mostly unfold linearly along a single timeline. This dimension aims to test the model's ability to synchronously track the independent state evolution of multiple interacting entities within the same time window.

\textbf{(3) Temporal Lags and Multi-Hop Dependencies.} Causal chains in real narratives often span multiple paragraphs. This dimension tests the model's ability to concatenate and reconstruct scattered event fragments with relative time offsets from across the text into a complete long-range causal chain.

\textbf{(4) Counterfactual Cascading Deduction.} This dimension tests the model's understanding of a system's intrinsic rules. By reversing or removing key historical nodes, the model must deduce subsequent timeline evolution based on the quantitative rules provided in the text.

\textbf{(5) Lagged Amplification of Minute Perturbations.} Inspired by the bullwhip effect in supply chains, this dimension aims to test the model's ability to capture long-range lag propagation and asymmetric causal relationships.

TempoCausal-Bench devises rigorous construction logics and standardized question templates across five core dimensions, as presented in Table \ref{tab:5}.

\begin{table}[htbp]
\centering
\small 
\caption{Construction of the five-dimensional test for TempoCausal-Bench.}
\label{tab:5}
\renewcommand{\arraystretch}{1.8} 
\begin{tabular}{m{0.18\textwidth} m{0.35\textwidth} m{0.4\textwidth}}
\toprule
Dimension & Core Tested Capability & Standard Question Template \\
\midrule
D1: Implicit Temporal Windows and Conditional Filtering & 
Guide the model away from explicit date matching; require it to delineate boundaries through event anchors and complete conditional filtering. & 
"During the period from [Event A's occurrence] to [Event B's occurrence], how many [specific entities/states] satisfy [Condition C]?" \\

D2: Concurrent Multi-Entity State Tracking & 
Break the single-entity linear narrative structure; synchronously track the states of multiple entities and identify logical conflicts. & 
"Throughout the entire period when [Entity X] is in [continuous State A], what complete [State B] transitions does [Entity Y] undergo?" \\

D3: Temporal Lags and Multi-Hop Dependencies & 
Test the model's ability to assemble scattered event fragments into a DAG, examining its topological reasoning level in reconstructing long-range causal chains across paragraphs. & 
"If the most critical precondition leading to [final Event Z] is [Event Y], then at the exact moment of [Event Y]'s occurrence (e.g., [time offset] prior), what action is [Entity X] performing?" \\

D4: Counterfactual Cascading Deduction & 
Examine the model's understanding of the text's intrinsic rules and require it to infer a counterfactual timeline based on those rules. & 
"Had [historical turning point Event A] not occurred, according to [the text's established rules/values], how would [Event B], originally scheduled at [Time T], have evolved?" \\

D5: Lagged Amplification of Minute Perturbations & 
Extract asymmetric causality and long-lag effects; identify early weak triggering signals. & 
"What was the most fundamental local triggering event that caused [Entity X] to make [a major structural change] at [late Time $T_2$]? From the first appearance of this weak signal to the eventual materialization of the macroscopic change, how long was the incubation period?" \\
\bottomrule
\end{tabular}
\end{table}

\subsection{Benchmark Construction Pipeline}
The benchmark construction adopts a three-stage pipeline to ensure logical rigor and data reliability. Prompt examples are provided in Tables \ref{tab:6}-\ref{tab:8}.

\textbf{(1) Initial Generation Stage.} The original Story from TimeQA-Hard is fed into an LLM, and combined with the five-dimensional specifications and construction logic, question-answer pairs corresponding to each dimension are generated in batches.

\textbf{(2) Correctness Verification.} GPT-5.4 and DeepSeek-V3.2 are used as verifiers to review the generated results, evaluating the correctness of answers and determining whether the questions are out of scope. When both LLMs simultaneously judge an answer as incorrect, GPT-5.4 is invoked to correct the answer. If both simultaneously judge the question as out of scope, the answer of that sample is marked as Unknown. If the two models disagree, the case is referred to human arbitration and corrected manually. According to statistics, the preliminary accuracy of the question-answer pairs constructed in this stage is 77.21\%, and another 19.53\% of the samples are marked as Unknown due to being out of scope.

\textbf{(3) Sample Secondary Reconstruction.} Samples marked as Unknown are reconstructed, with the model constrained to regenerate questions based solely on the events described in the original text. Subsequently, the newly generated samples are again passed through the aforementioned correctness verification process, and those that fail are manually corrected. After inspection, the construction accuracy of the question-answer pairs in this stage reaches 98.2\%, with the remaining 1.8\% of samples manually corrected, ultimately ensuring the reliability of the question-answer pairs.

Based on the above pipeline, we reconstruct 6,000 complex reasoning samples from 1,500 samples in a subset of TimeQA-Hard and split them into training and test sets at a ratio of 4:1. During dataset splitting, all variants generated from the same original sample are guaranteed to belong to the same set, thereby preventing data leakage at the source.

\definecolor{promptBlue}{RGB}{74,114,204}
\definecolor{promptBg}{RGB}{248,250,255}

\begingroup
\small 
\renewcommand{\arraystretch}{1.4} 
\arrayrulecolor{promptBlue} 

\begin{longtable}{|p{\dimexpr\textwidth-2\tabcolsep-2\arrayrulewidth\relax}|}
\caption{Question-answer pair construction prompt example.}
\label{tab:6} \\
\hline
\rowcolor{promptBlue}
\textcolor{white}{\textbf{Stage 1: Question-Answer Pairs Construction Prompt Example}} \\
\hline
\endfirsthead

\caption[]{Question-answer pair construction prompt example. (Continued)} \\
\hline
\rowcolor{promptBlue}
\textcolor{white}{\textbf{Stage 1: Question-Answer Pairs Construction Prompt Example}} \\
\hline
\endhead

\hline
\endfoot

\hline
\endlastfoot


\rowcolor{promptBg}
\textbf{Please generate 5 complex questions based on the provided story, following the methods outlined below. Provide the standard answer for each question. Do NOT provide explanations.} \\[0.8em]

\rowcolor{promptBg}
\textbf{Requirements for questions and answers:} \newline
- Use the original text content as much as possible for both questions and answers. \newline
- The generated questions will be used to test other models (evaluated by EM and F1 metrics), so the answers MUST be concise. Do not ask questions that require long textual answers. \\[0.8em]

\rowcolor{promptBg}
\textbf{Construction Methods:} \newline
\textbf{Dimension 1: Implicit Temporal Windows and Conditional Filtering} \newline
Core Testing Capability: Reject direct date matching. Require the model to define time boundaries via ``event anchors'' (Interval Bounding) and perform extraction and counting of specific entities or actions within those implicit boundaries. \newline
Construction Steps: \newline
Step 1: Select two non-simultaneous causal or sequential events from the text (Event A as the start, Event B as the end). \newline
Step 2: Specify a state or entity type that changes within this interval. \newline
Step 3: Add a distractor condition to be excluded (e.g., a person from the previous term, goods that were not successfully delivered). \newline
Question Template: ``Between [the occurrence of Event A] and [the occurrence of Event B], how many [specific entities/states] satisfy [Condition C]?'' \\[0.8em]

\rowcolor{promptBg}
\textbf{Dimension 2: Concurrent Multi-Entity State Tracking} \newline
Core Testing Capability: Break away from single-entity linear tracking. Require the model to simultaneously track the independent states of two or more entities within the same time slice or interval, and find their intersections or logical conflicts. \newline
Construction Steps: \newline
Step 1: Identify two or more interacting entities in the text (e.g., a politician and their party, competitors, or a company and a regulator). \newline
Step 2: Lock onto a persistent state of one entity (e.g., ``during the period of being out of office'' or ``while holding a certain position''). \newline
Step 3: Ask what key state transition the other entity underwent during this period. \newline
Question Template: ``Throughout the entire period when [Entity X] was in [persistent state A], what complete [state B] transitions did [Entity Y] experience?'' \\[0.8em]

\rowcolor{promptBg}
\textbf{Dimension 3: Temporal Lags and Multi-Hop Dependencies} \newline
Core Testing Capability: Test the model's ability to assemble independent events scattered across multiple paragraphs—with relative time offsets (e.g., ``two years later,'' ``subsequently,'' ``upon the expiration of X's term'')—into a complete directed acyclic graph (DAG). \newline
Construction Steps: \newline
Step 1: Locate an outcome event in the text. \newline
Step 2: Trace backward to identify multiple preceding steps leading to that event, ensuring these steps are not adjacent in the text but described with time offsets. \newline
Step 3: Ask about a hidden node somewhere in the middle of this chain. \newline
Question Template: ``If the most critical prerequisite leading to [final event Z] is [event Y], then exactly [time offset, e.g., two years earlier / the previous term] when [event Y] occurred, what action was [Entity X] performing?'' \\[0.8em]

\rowcolor{promptBg}
\textbf{Dimension 4: Counterfactual Cascading Deduction} \newline
Core Testing Capability: Examine the model's understanding of the text's internal logical rules. Remove or invert an already occurred historical node, and require the model to deduce the collapsed timeline outcome based on rules provided in the text (e.g., electoral seat mathematics, eligibility rules for holding office). \newline
Construction Steps: \newline
Step 1: Find a historical turning point in the text that involves explicit quantitative rules or prior qualification constraints (e.g., a very close vote, a narrow win in a bid). \newline
Step 2: Assume that this turning point did NOT happen (i.e., invert the variable). \newline
Step 3: Ask how the immediate subsequent historical event on the counterfactual timeline would be rewritten, without making the question too complex. \newline
Question Template: ``If [historical turning point event A] had not happened, given [the established rules/numbers in the text], what would have happened to [event B], which originally occurred at [time T]?'' \\[0.8em]

\rowcolor{promptBg}
\textbf{Dimension 5: Lagged Amplification of Minute Perturbations} \newline
Core Testing Capability: Extract asymmetric causality and long temporal lags. The model must identify, within a grand narrative, a seemingly trivial local event early on that—after a long delay—becomes the root cause of a major global change. \newline
Construction Steps: \newline
Step 1: Find a major structural change at the end or later part of the text (e.g., complete withdrawal from politics, company bankruptcy, system overhaul). \newline
Step 2: Locate the earliest trigger in the text that led to this outcome (e.g., a minor scandal involving a subordinate years ago, an early slight mistake). \newline
Step 3: Establish this lagged causal link and require the model to calculate the time span from the emergence of the ``minor disturbance'' to the occurrence of the ``global collapse.'' \newline
Question Template: ``When did the most fundamental local trigger that led [Entity X] to make [major structural change] at [late time $T_2$] occur? From the first appearance of that minor signal to the eventual macro-level change, how long was the incubation period?'' \\[0.8em]

\rowcolor{promptBg}
\textbf{Story:} \newline
\{Economic and Financial Committee (European Union). The Economic and Financial Committee (EFC) is an advisory body of the European Union, as defined by Article 134 of the Treaty on the Functioning of the European Union. Its president is also the president of the Eurogroup Working Group, which prepares dossiers for approval by the Eurogroup, whose decisions are generally ratified by ECOFIN. History: Before the third stage of the Economic and Monetary Union (EMU), which began on 1 January 1999, the Economic and Financial Committee was preceded by the Monetary Committee, which was established by Article 105 of the Treaty of Rome in 1957. In May, as part of the European Semester 2016 Spring Package, the committee was tasked with making reports on Belgium, Finland, and Italy's government debt. Agenda and composition of the EFC...\} \\[0.8em]

\rowcolor{promptBg}
\textbf{Please return the output in JSON format only. The root must be a JSON object with a key ``qa\_list'', and its value is a list of 5 objects. Each object should have the following keys:} \newline
- ``dimension'': The dimension number (e.g., ``Dimension 1'') \newline
- ``question'': The generated question \newline
- ``answer'': The concise standard answer \\

\end{longtable}
\endgroup

\begingroup
\small 
\renewcommand{\arraystretch}{1.4} 
\arrayrulecolor{promptBlue} 

\begin{longtable}{|p{\dimexpr\textwidth-2\tabcolsep-2\arrayrulewidth\relax}|}
\caption{Question-answer pair checking prompt example.}
\label{tab:7} \\
\hline
\rowcolor{promptBlue}
\textcolor{white}{\textbf{Stage 2: Question-Answer Pair Checking Prompt Example}} \\
\hline
\endfirsthead

\caption[]{Question-answer pair checking prompt example. (Continued)} \\
\hline
\rowcolor{promptBlue}
\textcolor{white}{\textbf{Stage 2: Question-Answer Pair Checking Prompt Example}} \\
\hline
\endhead

\hline
\endfoot

\hline
\endlastfoot


\rowcolor{promptBg}
\textbf{Please evaluate the following question-answer pairs based on the provided story.} \newline
For each question, check if the provided answer is correct according to the story. \newline
If the answer is incorrect, please correct it. \newline
If the question is out of the scope of the story and cannot be answered, set the corrected answer to ``Unknown''. \\[0.8em]

\rowcolor{promptBg}
\textbf{Requirements for the corrected answers:} \newline
- The generated questions will be used to test other models (evaluated by EM and F1 metrics), so the answers MUST be concise. Do not ask questions that require long textual answers. \\[0.8em]

\rowcolor{promptBg}
\textbf{Story:} \newline
\{Economic and Financial Committee (European Union). The Economic and Financial Committee (EFC) is an advisory body of the European Union, as defined by Article 134 of the Treaty on the Functioning of the European Union. Its president is also the president of the Eurogroup Working Group, which prepares dossiers for approval by the Eurogroup, whose decisions are generally ratified by ECOFIN. History: Before the third stage of the Economic and Monetary Union (EMU), which began on 1 January 1999, the Economic and Financial Committee was preceded by the Monetary Committee, which was established by Article 105 of the Treaty of Rome in 1957. In May, as part of the European Semester 2016 Spring Package, the committee was tasked with making reports on Belgium, Finland, and Italy's government debt. Agenda and composition of the EFC...\} \\[0.8em]

\rowcolor{promptBg}
\textbf{QA Pairs:} \newline
\{qa\_json\_str\} \\[0.8em]

\rowcolor{promptBg}
\textbf{Please return the output in JSON format only. The root must be a JSON object with a key ``checked\_qa\_list'', and its value is a list of objects corresponding to the input QA pairs in the same order. Each object should have the following keys:} \newline
- ``dimension'': The original dimension \newline
- ``question'': The original question \newline
- ``original\_answer'': The original answer \newline
- ``is\_correct'': true or false, whether the original answer was correct \newline
- ``answer'': The concise standard answer (use the original if it was correct, the corrected one if incorrect, or ``Unknown'' if out of scope) \\

\end{longtable}
\endgroup

\begingroup
\small 
\renewcommand{\arraystretch}{1.4} 
\arrayrulecolor{promptBlue} 

\begin{longtable}{|p{\dimexpr\textwidth-2\tabcolsep-2\arrayrulewidth\relax}|}
\caption{Sample reconstruction prompt example.}
\label{tab:8} \\
\hline
\rowcolor{promptBlue}
\textcolor{white}{\textbf{Stage 3: Sample Reconstruction Prompt Example}} \\
\hline
\endfirsthead

\caption[]{Sample reconstruction prompt example. (Continued)} \\
\hline
\rowcolor{promptBlue}
\textcolor{white}{\textbf{Stage 3: Sample Reconstruction Prompt Example}} \\
\hline
\endhead

\hline
\endfoot

\hline
\endlastfoot


\rowcolor{promptBg}
\textbf{You are a professional assistant for constructing and correcting datasets.} \newline
You need to strictly follow the specified dimensions and rules to rewrite the questions that cannot be answered (Unknown) based on the given original story (Story). Please ensure to return in the legal JSON format. \\[0.8em]

\rowcolor{promptBg}
\textbf{Dimension Construction Rules:} \newline
\{dimension\_rules\} \\[0.8em]

\rowcolor{promptBg}
\textbf{Original Story:} \newline
\{Economic and Financial Committee (European Union). The Economic and Financial Committee (EFC) is an advisory body of the European Union, as defined by Article 134 of the Treaty on the Functioning of the European Union. Its president is also the president of the Eurogroup Working Group, which prepares dossiers for approval by the Eurogroup, whose decisions are generally ratified by ECOFIN. History: Before the third stage of the Economic and Monetary Union (EMU), which began on 1 January 1999, the Economic and Financial Committee was preceded by the Monetary Committee, which was established by Article 105 of the Treaty of Rome in 1957. In May, as part of the European Semester 2016 Spring Package, the committee was tasked with making reports on Belgium, Finland, and Italy's government debt. Agenda and composition of the EFC...\} \\[0.8em]

\rowcolor{promptBg}
\textbf{Pending Q\&A Pairs:} \newline
\{qa\_json\_str\} \\[0.8em]

\rowcolor{promptBg}
\textbf{Task Requirements:} \newline
1. Carefully read the 5 pairs of questions and answers provided above. \newline
2. For the pairs where the ``answer'' is ``Unknown'', explain that the original question cannot be answered within the story (beyond the scope of the text). Please, based on the original story and the corresponding dimension rules, rewrite a new and challenging question and provide a concise answer. \newline
3. The rewritten question must be able to find a precise answer within the original text. It must strictly adhere to the ``core testing ability'' and ``question template'' in the corresponding dimension rules. \newline
4. For the pairs where the ``answer'' is not ``Unknown'', please keep the original text unchanged. \newline
Please return the result in a strict JSON format. The root node must be an array named ``rewritten\_qa\_list'', and this array must contain 5 objects, maintaining the same order as the input question-answer pairs. \\[0.8em]

\rowcolor{promptBg}
\textbf{The format of each object is as follows:} \newline
\{ \newline
``dimension'': ``The name of the dimension, for example, Dimension 5'', \newline
``question'': ``The original question (if not modified) or the new question after rewriting'', \newline
``answer'': ``The original answer (if not modified) or the concise answer after rewriting'', \newline
``is\_rewritten'': true or false (if rewriting has been done, it is true; otherwise, it is false) \newline
\} \\

\end{longtable}
\endgroup

\section{TCR-Former Training Strategy}
\label{app:TCR-Former Training Strategy}
To equip the model with end-to-end reasoning capability from fragmented narratives to topological causal reconstruction while avoiding gradient conflicts and representation degradation commonly encountered in multi-task joint optimization, we propose a three-stage training strategy. During training, the optimization objectives for temporal perception, cross-modal alignment, and causal reasoning are decoupled, enabling synergistic incremental improvement while ensuring stable convergence of each module. In the first stage, the temporal causal encoder is trained, with the base language model's parameters frozen and only the event anchor embeddings, time span biases, and the timestamp regression head updated. In the second stage, the cross-modal projector is trained, with all encoder and decoder parameters frozen and only the nonlinear projection layer optimized. In the third stage, end-to-end instruction fine-tuning is performed, with the encoder and projector frozen and the decoder base efficiently fine-tuned using LoRA. TCR-Former is trained on 4 × NVIDIA A100-40G GPUs.

\subsection{First Stage Temporal Causal Encoder Pretraining}
The training objective of the first stage is to establish the modeling capacity of the unified temporal causal encoder $E_\theta$ for implicit temporal dynamics and event structures. The encoder receives a sequence of original text chunks, extracts high-dimensional causal latent variables via learnable event anchors and time span biases, and predicts the physical temporal boundaries corresponding to each anchor. During training, all parameters of the base language model remain frozen, and only the event anchor embeddings, time bias parameters, and the timestamp regression head are updated, thereby reducing memory overhead and avoiding catastrophic forgetting. Three loss functions jointly provide the training signal.

\noindent\textbf{(1) Timestamp Regression Loss.} A lightweight regression head $h_\tau: \mathbb{R}^d \to \mathbb{R}$ is attached to the end of the encoder, mapping the hidden state $\mathbf{z}_k \in \mathbb{R}^d$ of the $k$-th query token to a normalized timestamp prediction $\hat{t}_k$. Given ground-truth time labels $\{t_k\}_{k=1}^M$, this loss applies mean squared error constraints as defined in Equation~\eqref{eq:10}.
\begin{equation}
\mathcal{L}_{\mathrm{time}} = \frac{1}{M} \sum_{k=1}^{M} \| h_\tau(\mathbf{z}_k) - t_k \|_2^2
\tag{10}
\label{eq:10}
\end{equation}
The timestamp regression loss drives the encoder to align event semantics with the physical timeline in the latent space, providing temporal priors for subsequent causal ordering.

\noindent\textbf{(2) Semantic Reconstruction Loss.} To prevent factual information decay during latent variable compression, we introduce a cosine similarity constraint, requiring the causal latent variables $\mathbf{Z}_{\mathrm{causal}}$ to retain the global semantic kernel of the original text tokens. Denote the global average pooling representations of the causal latent variables and text tokens in the batch $b$ as $\bar{\mathbf{z}}^{(b)}$ and $\bar{\mathbf{v}}^{(b)}$, respectively. The semantic reconstruction loss is then defined as Equation~\eqref{eq:11}.
\begin{equation}
\mathcal{L}_{\mathrm{sem}} = 1 - \frac{1}{B} \sum_{b=1}^{B} \frac{\langle \bar{\mathbf{z}}^{(b)}, \bar{\mathbf{v}}^{(b)} \rangle}{\| \bar{\mathbf{z}}^{(b)} \|_2 \cdot \| \bar{\mathbf{v}}^{(b)} \|_2}
\tag{11}
\label{eq:11}
\end{equation}
The semantic reconstruction loss ensures that the causal latent variables carry not only topological structure information but also necessary factual semantics, laying a foundation for cross-modal alignment in the second stage training.

\noindent\textbf{(3) Counterfactual Contrastive Loss.} To enable the encoder to learn to distinguish genuine causal dependencies from pure temporal co-occurrences, we construct a contrastive learning objective based on topological interventions. During training, a subset of samples in a batch is selected with probability $p_{\mathrm{cf}}$ for intervention augmentation. For each selected sample, a proportion $\rho_{\mathrm{cf}}$ of nodes are randomly sampled from their non-empty event anchors as the intervention set $k_\psi$. The encoder computes in parallel the factual latent representation $\mathbf{Z}_{\mathrm{factual}}$ and the counterfactual latent representation $\mathbf{Z}_{\mathrm{cf}}$ after severing all attention connections of the nodes in $k_\psi$. These two representations are required to maintain a Euclidean distance of no less than a preset margin $m$ in the feature space, explicitly modeling the representational shift induced by causal chain breakage. The computation is given in Equation~\eqref{eq:12}.
\begin{equation}
\mathcal{L}_{\mathrm{cf}} = \frac{1}{B} \sum_{b=1}^{B} \mathrm{ReLU} \left( m - \| \mathbf{Z}_{\mathrm{factual}}^{(b)} - \mathbf{Z}_{\mathrm{cf}}^{(b)} \|_2 \right)
\tag{12}
\label{eq:12}
\end{equation}
For unselected samples, $\mathcal{L}_{\mathrm{cf}}$ is set to zero. To prevent intervention noise from disrupting early representation learning, a curriculum schedule is introduced. Specifically, $\mathcal{L}_{\mathrm{cf}}$ is completely turned off during the first 30\% of training steps. For the 31\%--70\% steps, $p_{\mathrm{cf}} = 0.1$ and $\rho_{\mathrm{cf}} = 0.1$ are used, and for the final 30\% of steps, $p_{\mathrm{cf}} = 0.2$ and $\rho_{\mathrm{cf}} = 0.2$ are set.

\noindent\textbf{(4) First Stage Loss Function.} The overall optimization objective of the first stage is a weighted sum of the three losses, as given in Equation~\eqref{eq:13}.
\begin{equation}
\mathcal{L}_{\mathrm{phase1}} = \alpha \mathcal{L}_{\mathrm{time}} + \beta \mathcal{L}_{\mathrm{sem}} + \gamma \mathcal{L}_{\mathrm{cf}}
\tag{13}
\label{eq:13}
\end{equation}
where $\alpha$, $\beta$, $\gamma$ are weighting coefficients. Considering the importance of each term, we set $\alpha = 1.0$, $\beta = 0.5$, and $\gamma = 0.2$.

\subsection{Second Stage Cross-Modal Projection Alignment}
The second stage focuses on training the nonlinear projector $\mathcal{P}: \mathbb{R}^{d_{\mathrm{enc}}} \to \mathbb{R}^{d_{\mathrm{dec}}}$, whose function is to map the causal latent variables $\mathbf{Z}_{\mathrm{causal}}$ output by the encoder into the semantic space of the LLM decoder. In this stage, all parameters of both the encoder and decoder base models are frozen, and only the projector module is trained. The training follows the standard autoregressive language modeling objective, and a bypass-aware label masking strategy is introduced.

The decoder input sequence $\mathbf{E}_{\mathrm{input}}$ is concatenated in the order of $[\mathbf{V}_{\mathrm{text}}; \mathcal{P}(\mathbf{Z}_{\mathrm{causal}}); \mathbf{E}_{\mathrm{prompt}}]$. When computing the cross-entropy loss, supervision is only applied to the target response portion following the prompt text, and the conditional prefix composed of $\mathbf{V}_{\mathrm{text}}$ and $\mathcal{P}(\mathbf{Z}_{\mathrm{causal}})$ does not participate in the loss calculation. Formally, let $T_{\mathrm{target}}$ be the set of token indices of the target response after the prompt text. The language modeling loss is defined as Equation~\eqref{eq:14}.
\begin{equation}
\mathcal{L}_{LM} = - \sum_{t \in T_{\mathrm{target}}} \log p_\theta (y_t | \mathbf{E}_{\mathrm{input}}, y_{<t})
\tag{14}
\label{eq:14}
\end{equation}
This masking strategy ensures that the projector focuses on transforming the causal latent variables into structured instructions interpretable by the LLM.

\vspace{1em}

\subsection{Third Stage End-to-End Instruction Tuning}

The third stage performs end-to-end instruction tuning across the entire TCR-Former architecture, strengthening its reasoning capabilities for complex temporal question-answering and counterfactual deduction tasks. In this stage, all encoder and projector parameters are frozen, and low-rank adaptation (LoRA) \cite{R64} is used for parameter-efficient fine-tuning of the decoder base. The rank hyperparameter $r$ of the low-rank adaptation matrices $\Delta\mathbf{W} = \mathbf{A}\mathbf{B}$ is set to 64, and the scaling factor $\alpha$ is set to 128. The adaptation modules are applied to the attention projection layers and the linear layers in the feed-forward networks. This strategy efficiently injects temporal causal reasoning priors while preserving the base model's general language capabilities.

\vspace{1em}

\subsection{Training Dataset Construction}
To support the three-stage training of TCR-Former, we constructed a structured training dataset based on multiple public temporal reasoning benchmarks and the TempoCausal-Bench.

\subsubsection{First Stage Dataset}
The first stage aims to train the encoder for temporal causal perception and structured modeling. We extracted and integrated 60,000 long-text narrative paragraphs from public datasets, including TGQA \cite{R4}, TempReason \cite{R58}, and TimeQA \cite{R8}. These texts cover multiple domains, including biographies, historical events, and organizational changes. The core of data construction lies in extracting event sequences that contain causal associations from the text along with their corresponding times. Specifically, for each text, we utilized DeepSeek-V3.2 to annotate explicit and implicit temporal and causal cues, forming a set of ordered event-timestamp pairs. Each event is summarized in a concise description and paired with the time point at which it occurs. The physical time is first uniformly converted to a floating-point number with years as the unit. For time points specified only with a year, the fractional part is zero. Subsequently, all timestamps are normalized to the [0, 1] interval, enabling the encoder's regression head to predict relative temporal positions stably. Through this process, each training sample contains a sequence of original text chunks, a sequence of extracted causal event descriptions, and a corresponding sequence of normalized target timestamps. Table \ref{tab:9} provides a simplified example of a training sample.

\begingroup
\small 
\renewcommand{\arraystretch}{1.4} 
\arrayrulecolor{promptBlue} 

\begin{longtable}{|p{\dimexpr\textwidth-2\tabcolsep-2\arrayrulewidth\relax}|}
\caption{Examples of samples from the first stage dataset.}
\label{tab:9} \\
\hline
\rowcolor{promptBlue}
\textcolor{white}{\textbf{Examples of Samples From the First Stage Dataset}} \\
\hline
\endfirsthead

\caption[]{Examples of samples from the first stage dataset. (Continued)} \\
\hline
\rowcolor{promptBlue}
\textcolor{white}{\textbf{Examples of Samples From the First Stage Dataset}} \\
\hline
\endhead

\hline
\endfoot

\hline
\endlastfoot


\rowcolor{promptBg}
\{ \newline
\hspace*{1em}\textbf{"text\_chunks"}: [ \newline
\hspace*{2em}"Jaroslav Pelikan ... was born on December 17, 1923, in Akron, Ohio ... He earned a seminary degree from Concordia Seminary in 1946 ... He completed his Ph.D. at the University of Chicago in 1953. Subsequently, in 1962, he joined the faculty at Yale University ..." \newline
\hspace*{1em}], \newline
\hspace*{1em}\textbf{"prompt\_texts"}: [ \newline
\hspace*{2em}"Extract the chronological timeline of causal events." \newline
\hspace*{1em}], \newline
\hspace*{1em}\textbf{"target\_events"}: [ \newline
\hspace*{2em}"Jaroslav Pelikan earned a degree from Concordia Seminary.", \newline
\hspace*{2em}"Jaroslav Pelikan completes Ph.D. at the University of Chicago.", \newline
\hspace*{2em}"Jaroslav Pelikan joins the faculty at Yale University." \newline
\hspace*{1em}], \newline
\hspace*{1em}\textbf{"target\_timestamps"}: [0.35, 0.52, 0.78] \newline
\} \\

\end{longtable}
\endgroup

\subsubsection{Second Stage Dataset}
The second stage focuses on training the projector to align the causal latent variables with the LLM's semantic space. This dataset shares the same textual sources as the first stage but is constructed in a form that serves the autoregressive language modeling objective. From 40,000 texts, we used DeepSeek-V3.2 to organize further the causal event sequences generated in the first stage into a coherent natural-language passage. A training sample consists of the original text chunks, an instruction prompt that triggers the decoder's behavior, and the structured timeline text serving as the generation target. In this stage, the projector loss is applied only to the target part's token sequence, enabling the projector to learn to convert the fixed-length causal latent-variable sequence output by the encoder into a dynamic-length semantic representation that the decoder can parse. Table \ref{tab:10} provides a simplified example of a training sample.

\begingroup
\small 
\renewcommand{\arraystretch}{1.4} 
\arrayrulecolor{promptBlue} 

\begin{longtable}{|p{\dimexpr\textwidth-2\tabcolsep-2\arrayrulewidth\relax}|}
\caption{Examples of samples from the second stage dataset.}
\label{tab:10} \\
\hline
\rowcolor{promptBlue}
\textcolor{white}{\textbf{Examples of Samples From the Second Stage Dataset}} \\
\hline
\endfirsthead

\caption[]{Examples of samples from the second stage dataset. (Continued)} \\
\hline
\rowcolor{promptBlue}
\textcolor{white}{\textbf{Examples of Samples From the Second Stage Dataset}} \\
\hline
\endhead

\hline
\endfoot

\hline
\endlastfoot


\rowcolor{promptBg}
\{ \newline
\hspace*{1em}\textbf{"text\_chunks"}: [ \newline
\hspace*{2em}"Jaroslav Pelikan ... was born on December 17, 1923, in Akron, Ohio ... He earned a seminary degree from Concordia Seminary in 1946 ... He completed his Ph.D. at the University of Chicago in 1953. Subsequently, in 1962, he joined the faculty at Yale University ..." \newline
\hspace*{1em}], \newline
\hspace*{1em}\textbf{"prompt\_texts"}: [ \newline
\hspace*{2em}"Please reconstruct the exact timeline of events based on the text." \newline
\hspace*{1em}], \newline
\hspace*{1em}\textbf{"target"}: "Jaroslav Pelikan earns his degree from Concordia Seminary around 1946.\textbackslash{}nJaroslav Pelikan completes his Ph.D. at the University of Chicago around 1953.\textbackslash{}nJaroslav Pelikan joins the faculty at Yale University around 1962." \newline
\} \\

\end{longtable}
\endgroup

\subsubsection{Third Stage Dataset}
The third stage dataset is used for end-to-end instruction tuning, aiming to finally unlock the model's reasoning capabilities for complex temporal causal question answering and counterfactual deduction. Based on the construction pipeline of the TempoCausal-Bench, we first utilized GPT-5.4 as the base model. We ran the TCR-Agent framework on the TempoCausal-Bench training set to generate high-quality question-answer pairs. Each pair contains a complex causal reasoning question and its structured CoT answer, which strictly follows the tag sequence of <checking> → <reasoning> → <reflection>. Through this approach, we obtained 4,000 instruction-tuning samples that embody deep causal topology and counterfactual deduction logic, providing a learnable reasoning paradigm for the model.

\section{LLM-Acc}
\label{app:LLM-Acc}
Since EM and F1 lack semantic judgment, we introduce a semantic-level accuracy metric, LLM-Acc. This metric employs two LLMs, DeepSeek-V3.2 and Gemini-3.1-flash, as independent evaluators. For a generated answer, each evaluator compares the predicted answer with the standard answer for semantic equivalence and gives a binary judgment of correct or incorrect. The final LLM-Acc is the average of the two evaluators’ scores. During evaluation, the generation temperature of both evaluators is set to 0 to ensure determinism and reproducibility. The prompt template for the LLM is shown in Table \ref{tab:11}.

\begingroup
\small 
\renewcommand{\arraystretch}{1.4} 
\arrayrulecolor{promptBlue} 

\begin{longtable}{|p{\dimexpr\textwidth-2\tabcolsep-2\arrayrulewidth\relax}|}
\caption{LLM judgment prompt example.}
\label{tab:11} \\
\hline
\rowcolor{promptBlue}
\textcolor{white}{\textbf{LLM Judgment Prompt Example}} \\
\hline
\endfirsthead

\caption[]{LLM judgment prompt example. (Continued)} \\
\hline
\rowcolor{promptBlue}
\textcolor{white}{\textbf{LLM Judgment Prompt Example}} \\
\hline
\endhead

\hline
\endfoot

\hline
\endlastfoot


\rowcolor{promptBg}
\textbf{You are an expert evaluator for temporal causal reasoning tasks.} \\[0.8em]

\rowcolor{promptBg}
Your job is to judge whether a model's predicted answer is semantically equivalent to the ground-truth answer. Do not focus on superficial wording differences. Focus on the core facts, logical consistency, and causal implications. If the prediction conveys the same meaning as the ground truth, it should be considered correct. \\[0.8em]

\rowcolor{promptBg}
\textbf{Question:} \{question\} \newline
\textbf{Ground-Truth Answer:} \{ground\_truth\} \newline
\textbf{Model's Response Answer:} \{model\_response\} \\[0.8em]

\rowcolor{promptBg}
Based on the above, is the Response Answer semantically correct? Please first provide a brief reasoning for your decision, then output your final verdict. Use the following format exactly: \\[0.8em]

\rowcolor{promptBg}
\textless{}reasoning\textgreater{} \newline
[Your reasoning here] \newline
\textless{}/reasoning\textgreater{} \newline
\textless{}verdict\textgreater{} \newline
Correct or Incorrect \newline
\textless{}/verdict\textgreater{} \\

\end{longtable}
\endgroup

\begin{figure}[t]
\centering
\includegraphics[width=14cm]{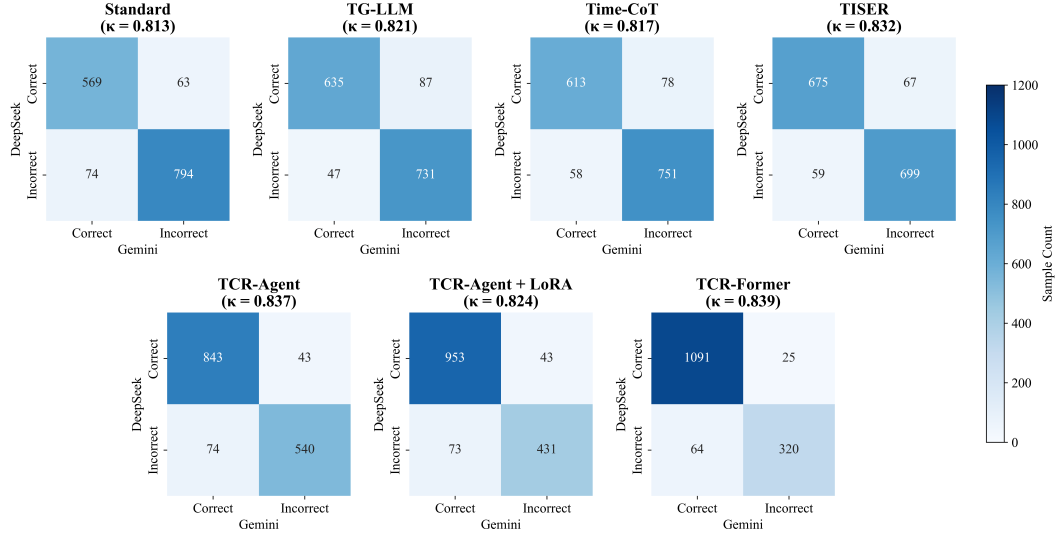}
\caption{Confusion matrix of the LLM judgment (Qwen-3-8B).}
\label{fig:4}
\end{figure}

Fig. \ref{fig:4} illustrates the confusion matrix of LLM judgments based on samples from Qwen-3-8B, and we use Cohen’s Kappa to measure the inter-judge agreement between the two LLMs. The average Kappa value across different methods is 0.826, indicating good agreement. To further verify the reliability of this automatic evaluation metric, we conducted a human review on 200 randomly sampled instances. The absolute deviation between human accuracy and LLM-Acc is 1.5\%, validating that the automatic evaluation scheme using LLMs as evaluators can effectively replace human evaluation with sound robustness.

\section{Baseline Methods}
\label{app:Baseline Methods}
\subsection{TG-LLM}
The CoT method in TG-LLM \cite{R4} aims to guide the LLM to perform rigorous temporal reasoning with the aid of explicit event context and external knowledge through a highly structured prompt design. As shown in Table \ref{tab:12}, in its specific prompt construction, the model input first contains a natural-language timeline that clearly defines the start and end times of all relevant events in a list format. Next is the temporal logic question to be answered. To compensate for the LLM’s deficiencies in mathematical and commonsense computation, this method explicitly introduces a Useful Information module in the prompt, directly injecting key external precomputed knowledge, including the chronological order of time points and the results of precise numerical difference calculations. After providing these enriched premises, the method employs “Let’s think step by step” as a trigger statement to regulate the model’s output of a complete reasoning path. Under this paradigm, the model can strictly follow a logical chain, first locating the specific time boundaries of the target event, then substituting the provided numerical values for span comparison, and finally deriving the conclusion based on the preceding explicit deduction steps.

\begingroup
\small 
\renewcommand{\arraystretch}{1.4} 
\arrayrulecolor{promptBlue} 

\begin{longtable}{|p{\dimexpr\textwidth-2\tabcolsep-2\arrayrulewidth\relax}|}
\caption{TG-LLM prompt example.}
\label{tab:12} \\
\hline
\rowcolor{promptBlue}
\textcolor{white}{\textbf{TG-LLM Prompt Example}} \\
\hline
\endfirsthead

\caption[]{TG-LLM prompt example. (Continued)} \\
\hline
\rowcolor{promptBlue}
\textcolor{white}{\textbf{TG-LLM Prompt Example}} \\
\hline
\endhead

\hline
\endfoot

\hline
\endlastfoot


\rowcolor{promptBg}
\textbf{Example:} \newline
\textbf{Timeline:} \newline
(Liam Mitchell was born in Harrison) starts at 1885 \newline
(Maddox Reynolds was born in Glenville) starts at 1893 \newline
(Liam Mitchell was married to Emma Scott) starts at 1916 \newline
(Emma Scott was married to Liam Mitchell) starts at 1916 \newline
(Liam Mitchell was married to Emma Scott) ends at 1918 \newline
(Emma Scott was married to Liam Mitchell) ends at 1918 \newline
(Maddox Reynolds was married to Liam Mitchell) starts at 1922 \newline
(Liam Mitchell was married to Maddox Reynolds) starts at 1922 \newline
(Maddox Reynolds was married to Liam Mitchell) ends at 1928 \newline
(Liam Mitchell was married to Maddox Reynolds) ends at 1928 \newline
(Liam Mitchell died in Boston) starts at 1941 \newline
(Emma Scott died in Oceanview) starts at 1984 \\[0.8em]

\rowcolor{promptBg}
\textbf{Question:} True or false: Event (Liam Mitchell was married to Maddox Reynolds) was longer in duration than event (Emma Scott was married to Liam Mitchell)? \\[0.8em]

\rowcolor{promptBg}
\textbf{Useful information:} \newline
1885 before 1893 before 1916 before 1918 before 1922 before 1928 before 1941 \newline
$1918 - 1916 = 2$ \newline
$1928 - 1893 = 35$ \newline
$1928 - 1922 = 6$ \newline
$1941 - 1918 = 23$ \newline
$2 < 6 < 23 < 35$ \\[0.8em]

\rowcolor{promptBg}
\textbf{Answer:} Let's think step by step. \newline
To determine whether the statement is true or false, we need to compare the durations of (Liam Mitchell was married to Maddox Reynolds) and (Emma Scott was married to Liam Mitchell). The duration for each event can be calculated as follows: \newline
(Liam Mitchell was married to Maddox Reynolds): \newline
(Liam Mitchell was married to Maddox Reynolds) starts at 1922 \newline
(Liam Mitchell was married to Maddox Reynolds) ends at 1928 \newline
$1928 - 1922 = 6$ \newline
(Emma Scott was married to Liam Mitchell): \newline
(Emma Scott was married to Liam Mitchell) starts at 1916 \newline
(Emma Scott was married to Liam Mitchell) ends at 1918 \newline
$1918 - 1916 = 2$ \newline
Now, we compare the duration of the two events: 6 is greater than 2; thus, the answer is True. \\[0.8em]

\rowcolor{promptBg}
\textbf{Test:} \newline
\textbf{Timeline:} \newline
(Chris Evans was born in Bristol, Connecticut) starts at 1948 \newline
(Oliver Jensen was born in Bristol, Connecticut) starts at 1948 \newline
(Chris Evans was married to Oliver Jensen) starts at 1970 \newline
(Oliver Jensen was married to Chris Evans) starts at 1970 \newline
(Chris Evans created Pearl Network) starts at 2005 \newline
(Chris Evans won prize Victory Achievement Award) starts at 2007 \newline
(Chris Evans was married to Oliver Jensen) ends at 2010 \newline
(Oliver Jensen was married to Chris Evans) ends at 2010 \newline
(Chris Evans created Pearl Network) ends at 2013 \\[0.8em]

\rowcolor{promptBg}
\textbf{Question:} True or false: Event (Chris Evans owned Pearl Network) was longer in duration than event (Oliver Jensen was married to Chris Evans)? \\[0.8em]

\rowcolor{promptBg}
\textbf{Useful information:} \newline
1948 before 1970 before 2005 before 2007 before 2010 \newline
$1970 - 1948 = 22$ \newline
$2007 - 1948 = 59$ \newline
$2007 - 2005 = 2$ \newline
$2010 - 1970 = 40$ \newline
$2 < 22 < 40 < 59$ \\[0.8em]

\rowcolor{promptBg}
\textbf{Answer:} Let's think step by step. \\

\end{longtable}
\endgroup

\subsection{Time-CoT}
Time-CoT \cite{R56} is a CoT method designed to enhance the ability of LLMs to handle factual question answering involving temporal reasoning. As shown in Table \ref{tab:13}, Time-CoT guides the model to perform step-by-step reasoning through a structured prompt template. First, the prompt requires the model to detail and describe the Standard Question Entity along the timeline dimension, and then focus on the specific Time Parameter for reasoning and matching. For example, when dealing with queries such as “What movie did Zoe Cassavetes direct in 2007 ?” or “Who was the quarterback of the New York Giants as of December 2010 ?”, Time-CoT forces the LLM first to construct the historical timeline of the entity, and then focus on the specific target time point to derive the exact final answer based on the intersection of the timeline.

\begingroup
\small 
\renewcommand{\arraystretch}{1.4} 
\arrayrulecolor{promptBlue} 

\begin{longtable}{|p{\dimexpr\textwidth-2\tabcolsep-2\arrayrulewidth\relax}|}
\caption{Time-CoT prompt example.}
\label{tab:13} \\
\hline
\rowcolor{promptBlue}
\textcolor{white}{\textbf{Time-CoT Prompt Example}} \\
\hline
\endfirsthead

\caption[]{Time-CoT prompt example. (Continued)} \\
\hline
\rowcolor{promptBlue}
\textcolor{white}{\textbf{Time-CoT Prompt Example}} \\
\hline
\endhead

\hline
\endfoot

\hline
\endlastfoot


\rowcolor{promptBg}
\textbf{Response Template:} \newline
Let's think step by step. First, we need to find out the \textless{}Standard Question Entity\textgreater{} according to the timeline. Then, we need to focus on the time \textless{}Explicit Time Parameter\textgreater{}. So we can obtain the answer is \textless{}Gold Answer\textgreater{}. Therefore, the answer(s) is/are ... \\[0.8em]

\rowcolor{promptBg}
\textbf{Example:} \newline
\textbf{Question:} Which movie did Zoe Cassavetes direct in 2007 ? \newline
\textbf{Answer:} Let's think step by step. First, we need to find out which movie did Zoe Cassavetes direct according to timeline. 2000: Men Make Women Crazy Theory, 2007: Broken English, 2012: The Powder Room, 2015: Day Out of Days...... Then, we need to focus on the time in 2007. So we can obtain the answer in Broken English. Therefore, the answer(s) is/are Broken English. \\[0.8em]

\rowcolor{promptBg}
\textbf{Question:} Who is the quarterback for the New York Giants as of December 2010 ? \newline
\textbf{Answer:} Let's think step by step. First, we need to find out who is the quarterback for the New York giants according to timeline. 1999-2003: Kerry Collins, 2003: Jesse Palmer, 2004: Kurt Warner, 2004-2019: Eli Manning, 2017: Geno Smith, 2019-2023: Daniel Jones, 2020: Colt McCoy, etc. Then, we need to focus on the time December 2010. So we can obtain the answer is Eli Manning. Therefore, the answer(s) is/are Eli Manning. \\[0.8em]

\rowcolor{promptBg}
\textbf{Test:} \newline
\textbf{Question:} Who was the president of the US when Adele Faccio was a teenager? \newline
\textbf{Answer:} Let's think step by step. \\

\end{longtable}
\endgroup

\subsection{TISER}
TISER \cite{R57} is a framework designed to enhance the temporal reasoning capabilities of large language models. It extends the traditional CoT into a four-stage cyclic inference paradigm comprising preliminary reasoning, timeline construction, reflection and verification, and answer generation, by scaling reasoning traces at test time. As shown in Table \ref{tab:14}, the specific workflow of TISER is as follows:

\textbf{(1) Stage 1 – Initial Reasoning.} The model first generates a preliminary, raw reasoning trajectory based on the provided question and temporal context.

\textbf{(2) Stage 2 – Timeline Construction.} Relevant temporal events are extracted from the context and the initial reasoning trajectory and organized into an explicit, ordered timeline.

\textbf{(3) Stage 3 – Reflection and Verification.} The model compares its initial reasoning trajectory with the newly constructed timeline to self-verify its logic. This stage is responsible for detecting and correcting inconsistencies, ambiguities, or omissions in the initial reasoning, thereby generating a refined reasoning trajectory.

\textbf{(4) Stage 4 – Answer Generation.} Relying on the error-corrected reasoning trajectory and the structured timeline, the model ultimately generates a concise, accurate, and logically coherent answer.

\begingroup
\small 
\renewcommand{\arraystretch}{1.4} 
\arrayrulecolor{promptBlue} 

\begin{longtable}{|p{\dimexpr\textwidth-2\tabcolsep-2\arrayrulewidth\relax}|}
\caption{TISER prompt example.}
\label{tab:14} \\
\hline
\rowcolor{promptBlue}
\textcolor{white}{\textbf{TISER Prompt Example}} \\
\hline
\endfirsthead

\caption[]{TISER prompt example. (Continued)} \\
\hline
\rowcolor{promptBlue}
\textcolor{white}{\textbf{TISER Prompt Example}} \\
\hline
\endhead

\hline
\endfoot

\hline
\endlastfoot


\rowcolor{promptBg}
\textbf{You are an AI assistant that uses a Chain of Thought (CoT) approach with reflection to answer queries. Follow these steps:} \newline
1. Reason through the problem step by step within the \textless{}reasoning\textgreater{} tags. \newline
2. Given your previous reasoning, identify relevant temporal events in the given context for answering the given question within \textless{}timeline\textgreater{} tags. Assume relations in the context are unidirectional. \newline
3. Reflect on your reasoning and the timeline to check for any errors or improvements within the \textless{}reflection\textgreater{} tags. \newline
4. Make any necessary adjustments based on your reflection. If additional reasoning is required, go back to Step 1 (reason through the problem step by step); otherwise, move to the next step (Step 5). \newline
5. Provide your final, concise answer within the \textless{}answer\textgreater{} tags. If the answer is a number, just output the number, nothing else. Otherwise, output the entity or event without any additional comments. \\[0.8em]

\rowcolor{promptBg}
\textbf{Additional Instructions:} \newline
1. [IF during data generation] Perform your reasoning knowing that the answer is: X. \newline
2. The \textless{}reasoning\textgreater{}, \textless{}reflection\textgreater{}, and \textless{}timeline\textgreater{} sections are for your internal reasoning process. All the reflection and the timeline have to be contained inside the thinking section. \newline
3. Do not use enumerations or lists when writing; use plain text such as paragraphs. \newline
4. The response to the query must be entirely contained within the \textless{}answer\textgreater{} tags. \\[0.8em]

\rowcolor{promptBg}
\textbf{Response Format:} \newline
\textless{}reasoning\textgreater{} \newline
[Your step-by-step reasoning goes here. This is your internal thought process.] \newline
\textless{}timeline\textgreater{} [Relevant temporal events for answering the given question.] \textless{}/timeline\textgreater{} \newline
\textless{}reflection\textgreater{} [Your reflection on your reasoning, checking for errors or changes required.] \textless{}/reflection\textgreater{} \newline
[Any adjustments to your thinking based on your reflection.] \newline
\textless{}/reasoning\textgreater{} \newline
\textless{}answer\textgreater{} [Your final, concise answer to the query.] \textless{}/answer\textgreater{} \\

\end{longtable}
\endgroup

\section{Ablation Study Setup}
\label{app:Ablation Study Setup}
To systematically analyze the actual contributions of each design element in TCR-Agent and TCR-Former, we conduct component-level ablation on the two frameworks on the TempoCausal-Bench.

\subsection{TCR‑Agent Ablation Setup}
The complete TCR-Agent pipeline based on GPT-5.4 contains three key stages: causal graph extraction, counterfactual intervention, and dual-track reasoning. By removing critical components from each stage individually, we obtain the following ablation variants.

\noindent \textbf{w/o Scratchpad:} During the causal graph extraction stage, the LLM is required to directly output the raw graph $G_{\mathrm{raw}}$ with nodes and edges, without generating intermediate reasoning steps. The instruction regarding the scratchpad is removed from the extraction prompt, and the content of the scratchpad $S$ is completely excluded from subsequent reasoning.

\noindent \textbf{w/o DAG Instantiation:} The structured DAG instantiation step is skipped. The extracted $G_{\mathrm{raw}}$ is directly serialized into a textual description as a list of natural language edges, without performing multi-dimensional attribute projection, timestamp consistency verification, or topological sorting with the directed acyclic constraint, thereby degenerating into a flat relational chain.

\noindent \textbf{w/o Counterfactual Intervention:} When processing counterfactual-type queries, the graph intervention engine is deactivated, and the original causal graph $G_{\mathrm{raw}}$ is used for reasoning, meaning no intervention operation is performed on the causal structure.

\noindent \textbf{w/o Dual-Track Reasoning:} The dual-track reasoning is downgraded to a single-track mode. Only the serialized description of the structured causal graph $D(G')$ is provided to the LLM as context, while the original narrative text is excluded, thereby isolating the auxiliary effect of unstructured factual details on structured logical deduction.

\noindent \textbf{Standard:} All external symbolic modules of TCR-Agent are completely removed, and only the original text and question are directly fed to GPT-5.4, which is required to output an answer, serving as the upper-bound reference for the pure language model baseline.

\subsection{TCR-Former Ablation Setup}
The complete TCR-Former architecture introduces core designs, including the causal-topological attention mask, explicit time-span bias, and fact-logic bypass fusion, on top of the Qwen-3-8B decoder. The role of each mechanism is examined through the following structural degradations.

\noindent \textbf{w/o Causal-Topological Attention Mask:} The causal-topological attention mask is replaced with the standard strictly lower-triangular causal mask.

\noindent \textbf{w/o Time Span Bias:} The learnable time span bias embeddings are frozen and removed, causing the initial representation of query tokens to degrade to relying solely on the position-independent base anchor embeddings. Meanwhile, the time bias matrix is no longer added during attention computation.

\noindent \textbf{w/o Fact-Logic Bypass Fusion:} At the decoder input, the word embedding track $\mathbf{E}_{\mathrm{text}}$ of the original text is removed, and only the aligned causal latent variables $\mathbf{Z}_{\mathrm{aligned}}$ from the encoder and the instruction prompt embedding $\mathbf{E}_{\mathrm{prompt}}$ are concatenated and fed into the decoder. This setup forces the generation process to rely entirely on the topology representation compressed in the latent space to examine the degree of factual information retention during compression.

\noindent \textbf{Only Decoder:} The encoder and the projection layer are removed, retaining only the native Qwen-3-8B. Since it cannot execute the first two training stages, we perform instruction tuning on it using the high-quality causal question-answering data generated in the third stage.

\noindent \textbf{Only Encoder + Decoder:} Only Qwen-3-0.6B is adopted as the encoder and Qwen-3-8B as the decoder, with the projection layer between the two models, undergoing the full three-stage training pipeline of TCR-Former.

\noindent \textbf{Standard:} Qwen-3-8B directly performs reasoning on the original text and question, serving as the performance reference for the pure backbone model.


\end{document}